\documentclass[iicol]{sn-jnl}
\usepackage{epstopdf}
\usepackage{graphicx}
\usepackage{cite}
\usepackage{overpic}
\usepackage{bm}
\usepackage{multirow}
\usepackage{color}

\usepackage[sort&compress,numbers]{natbib}

\jyear{2021}%

\theoremstyle{thmstyleone}%
\theoremstyle{thmstyletwo}%

\theoremstyle{thmstylethree}%

\makeatletter
\renewcommand{\maketag@@@}[1]{\hbox{\m@th\normalsize\normalfont#1}}%
\makeatother

\begin{document}

\title[Article Title]{Causal Reasoning Meets Visual Representation Learning: A Prospective Study}

\author[1]{\fnm{Yang} \sur{Liu}}

\author[1]{\fnm{Yushen} \sur{Wei}}

\author[1]{\fnm{Hong} \sur{Yan}}

\author[1]{\fnm{Guanbin} \sur{Li}}

\author[1]{\fnm{Liang} \sur{Lin}}

\affil[1]{\orgdiv{School of Computer Science and Engineering}, \orgname{Sun Yat-sen University}, \orgaddress{\city{Guangzhou} \postcode{510006},  \country{China}}}

\abstract{Visual representation learning is ubiquitous in various real-world applications, including visual comprehension, video understanding, multi-modal analysis, human-computer interaction, and urban computing. Due to the emergence of huge amounts of multi-modal heterogeneous spatial/temporal/spatial-temporal data in the big data era, the lack of interpretability, robustness, and out-of-distribution generalization are becoming the challenges of the existing visual models. The majority of the existing methods tend to fit the original data/variable distributions and ignore the essential causal relations behind the multi-modal knowledge, which lacks unified guidance and analysis about why modern visual representation learning methods easily collapse into data bias and have limited generalization and cognitive abilities. Inspired by the strong inference ability of human-level agents, recent years have therefore witnessed great effort in developing causal reasoning paradigms to realize robust representation and model learning with good cognitive ability. In this paper, we conduct a comprehensive review of existing causal reasoning methods for visual representation learning, covering fundamental theories, models, and datasets. The limitations of current methods and datasets are also discussed. Moreover, we propose some prospective challenges, opportunities, and future research directions for benchmarking causal reasoning algorithms in visual representation learning. This paper aims to provide a comprehensive overview of this emerging field, attract attention, encourage discussions, bring to the forefront the urgency of developing novel causal reasoning methods, publicly available benchmarks, and consensus-building standards for reliable visual representation learning and related real-world applications more efficiently. }


\maketitle

\section{Introduction}\label{sec1}

\begin{figure*}
    \centering
    \includegraphics[scale=0.245]{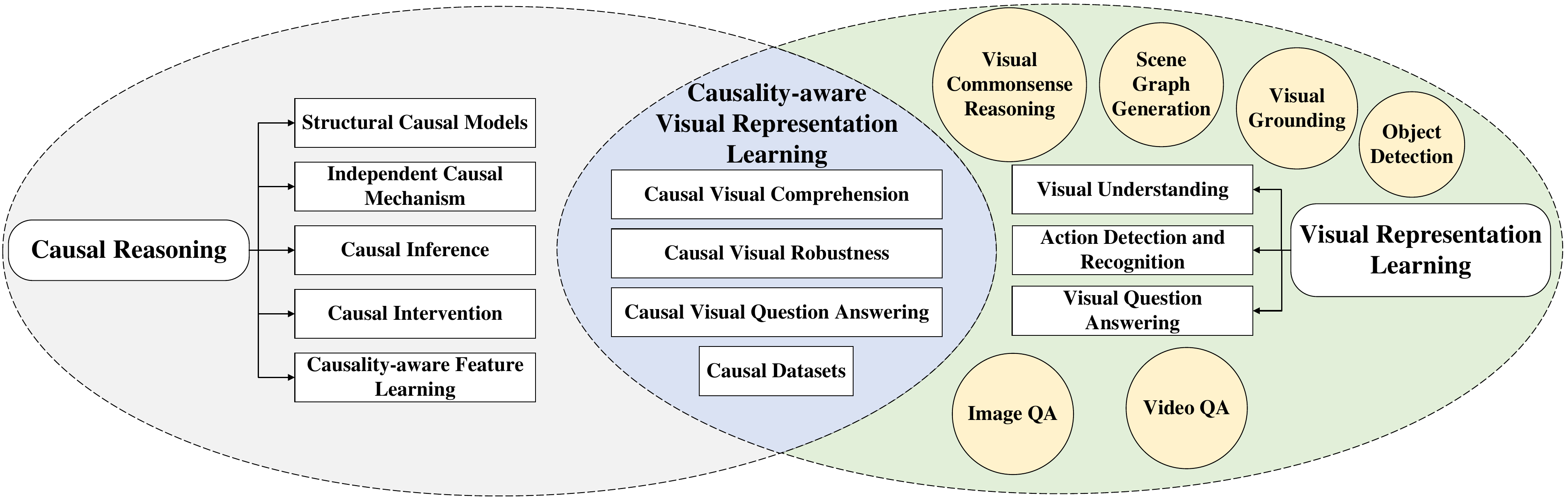}
    \caption{Overview of the structure of this paper, including the discussion of related methods, datasets, challenges, and the relations among causal reasoning, visual representation learning, and their integration.}
    \label{fig:framework}
\end{figure*}

With the emergence of huge amounts of heterogeneous multi-modal data, including images \cite{he2016deep,chen2020knowledge,akula2022cx}, videos \cite{wang2016temporal,zhou2018temporal,lin2019tsm,TCGL}, texts/languages \cite{busta2017deep,chen2021text,rastgoo2021sign}, audios \cite{gao2020listen,cheng2020look,chen2021distilling,AVCL}, and multi-sensor \cite{liu2018hierarchically,liu2018transferable,liu2019deep,liu2021semantics,zhu2022hybrid} data, deep learning based methods have shown promising performance for various computer vision and machine learning tasks, for example, the visual comprehension \cite{li2017instance,liang2018look,yang2020relationship,zhang2021advances}, video understanding \cite{wang2021action,pang2020self,liu2018global,wu2021comprehensive}, visual-linguistic analysis \cite{goyal2017making,cao2021knowledge,cao2021linguistically}, and multi-modal fusion \cite{liu2016combining,wang2019generative,ni2021cross}, etc. However, the existing methods rely heavily upon fitting the data distributions and tend to capture the spurious correlations from different modalities, and thus fail to learn the essential causal relations behind the multi-modal knowledge that have a good generalization and cognitive abilities. Inspired by the fact that most of the data in computer vision society are independent and identically distributed (i.i.d), a substantial body of literature \cite{geirhos2018imagenet,shetty2019not,hendrycks2019benchmarking,azulay2019deep} adopted data augmentation, pre-training, self-supervision, and novel architectures to improve the robustness of the state-of-the-art deep neural network architectures. However, it has been argued that such strategies only learn correlation-based patterns (statistical dependencies) from data and may not generalize well without the guarantee of the i.i.d setting \cite{peters2017elements}.

Due to the powerful ability of to uncover the underlying structural knowledge about data generating processes that allow interventions and generalize well across different tasks and environments, causal reasoning \cite{pearl2009causality,scholkopf2021toward,cheng2022evaluation,liu2022causal} offers a promising alternative to correlation learning. Recently, causal reasoning has attracted increasing attention in a myriad of high-impact domains of computer vision and machine learning, such as interpretable deep learning \cite{zhang2018visual,zhang2018interpretable,zhang2019interpreting,zhang2020extraction,zhang2020mining,zhang2020interpretable}, causal feature selection \cite{yu2020causality,yu2021unified,yu2022causal,guo2022error,li2021confounder,yu2021multilabel,yang2021causalvae,yang2021towards,li2021causality,ling2020using,yu2019multi,wu2019accurate,yu2019learning}, visual comprehension \cite{wang2021causal,yue2021counterfactual,huang2021deconfounded,zhang2021acre,zhang2020causal,tang2020unbiased,tang2020long,wang2020visual,chen2019counterfactual,shi2019explainable,chen2023visual}, visual robustness \cite{tang2021adversarial,hu2021distilling,yue2021transporting,yue2020interventional,yang2021learning,christiansen2021causal,mao2021generative,kyono2021exploiting}, visual question answering \cite{wu2017temporal,niu2021introspective,niu2021counterfactual,yang2021causal,qi2020two,chen2020counterfactual}, and video understanding \cite{wu2021learning,shi2021temporal,zhang2021learning,xu2021unsupervised,fire2017inferring,gangapure2017superpixel,xiong2016robot,liu2021temporal_2}. A common challenge of these causal methods is how to build a strong cognitive model that can fully discover causality and spatial-temporal relations.

In this paper, we aim to provide a comprehensive overview of causal reasoning for visual representation learning, attract attention, encourage discussions, and bring to the forefront the urgency of developing novel causality-guided visual representation learning methods. Although there are some surveys \cite{scholkopf2021toward,cheng2022evaluation,yang2021deconfounded,shen2021towards,yu2021unified,chen2020bias} about causal reasoning, these works are intended for general representation learning tasks such as deconfounding, out-of-distribution (OOD) generalization, and debiasing. Differently, our paper focuses on the systematic and comprehensive survey of related works, datasets, insights, future challenges and opportunities for causal reasoning, visual representation learning, and their integration. To present the review more concisely and clearly, this paper selects and cites related work by considering their sources, publication years, impact, and the cover of different aspects of the topic surveyed in this paper. The overview of the structure of this paper is shown in Fig. \ref{fig:framework}. Overall, the main contributions of this paper are given as follows.

Firstly, this paper presents the basic concepts of causality, the structural causal model (SCM), the independent causal mechanism (ICM) principle, causal inference, and causal intervention. Then, based on the analysis, this paper further gives some directions for conducting causal reasoning on visual representation learning tasks. Note that to the best of our knowledge, this paper is the first that proposes the potential research directions for causal visual representation learning.

Secondly, a prospective review is introduced to systematically and structurally review the existing works according to their efforts in the above-pointed directions for conducting causal visual representation learning more efficiently. We focus on the relation between visual representation learning and causal reasoning and provide a better understanding of why and how existing causal reasoning methods can be helpful in visual representation learning, as well as providing inspiration for future research and studies.

Thirdly, this paper explores and discusses future research areas and open problems related to using causal reasoning methods to tackle visual representation learning. This can encourage and support the broadening and deepening of research in the related fields.

The remainder of this paper is organized as follows. Section 2 provides the preliminaries, including the basic concepts of causality, the structural causal model (SCM), the independent causal mechanism (ICM) principle, causal inference, and causal intervention. Section 3 discusses the ways to use causal reasoning to learn robust features, which are the key techniques for visual representation learning. Section 4 reviews some recent visual learning tasks, including visual understanding, action detection and recognition, and visual question answering, including the discussions about the existing challenges of these visual learning methods. Section 5 reviews the related causality-based visual representation learning works systematically. Section 6 provides a review of existing causal datasets for visual learning. Section 7 proposes and discusses some future research directions and finally Section 8 gives the conclusions.

\section{Preliminaries}\label{sec2}
\subsection{Causal Learning and Reasoning}
As the sentence ``correlation is not causation" says, two variables are correlated does not mean that one of them causes the other. Actually, statistical learning models the correlations of data. By observing a sufficient amount of i.i.d. data, the statistical learning method can perform considerably well under i.i.d settings. However, when facing problems that do not satisfy i.i.d. assumptions, the performance of these methods often seems poor (e.g., image recognition models tend to predict ``bird" when seeing ``sky" in the image, since bird and sky usually appear simultaneously in the dataset). Causal learning \cite{pearl2009causality} is different from statistical learning, which aims to discover causal relationships beyond statistical relations. Learning causality requires machine learning methods not only to predict the outcome of i.i.d. experiments but also to reason from a causal perspective. Causal reasoning can be divided into three levels. The first level is association. The statistical machine learning methods mentioned above belong to this level. A typical question of association is ``How would the weather change when the sky is turning grey", which asks about the association between ``weather" and ``the appearance of the sky". The second level is intervention. An intervention-based question asks about the effect of the intervention (e.g., ``Would I become stronger if I go to the gym every day?''). Intervention-based questions require us to answer the outcome when taking specific treatment, which can not be answered by only learning data associations (e.g., If we only learn the associations, then if we observe that a man who goes to the gym every day may not be stronger than a professional athlete, we may conclude that going to the gym not always makes you stronger). The third level is counterfactual. A typical form of a counterfactual question is ``What if I had...", which focuses on the outcome when the condition is not realized. Counterfactual inference aims to compare different outcomes under the same condition, but the antecedent of the counterfactual question is not real.

\subsection{Structural Causal Models}
The structural causal model (SCM) considers the formulation of a causality style. Assume that we have a set of variables $X_1, X_2,...,X_n$, each variable is a vertex of a causal graph (i.e., a DAG describes causal relations of variables). Then, those variables could be written as the outcome of a function:
\begin{equation}
    X_i = f_i(\textbf{PA}_i, U_i) \approx P(X_i \vert \textbf{PA}_i)~~(i=1,...,n)
\end{equation}
where $\textbf{PA}_i$ indicates the parents of $X_i$ in the causal graph, and $U_i$ refers to unmeasured factors such as noise. The deterministic function gives a mathematical form of the effect of direct causes of $X_i$ on the variable $X_i$. Using the graphical causal model and SCM language, we can express joint distributions as follows:
\begin{equation}
    P(X_1, X_2,..., X_n) = \prod_{i=1}^n P(X_i \vert \textbf{PA}_i)
    \label{decomposition}
\end{equation}

Eq. \ref{decomposition} is called a product decomposition of the joint distributions. After the decomposition and graphical modeling, the causal relations and effects of a dataset can be represented as the causal graph and the joint distribution.

\subsection{Independent Causal Mechanism}
The independent causal mechanism principle \cite{scholkopf2021toward} can be expressed as follows:

\textit{\textbf{ICM principle}: The causal generative process of a system's variables is composed of autonomous modules that do not inform or influence each other. In the probabilistic case, this means that the conditional distribution of each variable given its causes (i.e., its mechanism) does not inform or influence the other conditional distributions.}

The ICM principle describes the independence of causal mechanisms. If we conceive that the real world is composed of modules in variable styles, then the modules could represent the physically independent mechanisms of the world. When applying the ICM principle to the disentangled factorization Eq. \ref{decomposition}, it can be written as \cite{scholkopf2021toward}:

(1) Changing (or performing an intervention upon) one
mechanism $P(X_i\vert \textbf{PA}_i)$ does not change any of the
other mechanisms $P(X_j \vert \textbf{PA}_j ) (i \neq j)$.

(2) Knowing some other mechanisms $P(X_i\vert \textbf{PA}_i)$
does not give us information about a mechanism
 $P(X_j \vert \textbf{PA}_j ) (i \neq j)$.

The ICM principle guarantees that our intervention on one mechanism does not affect others, which further reveals the possibility of transferring knowledge across domains that have the same modules.
\subsection{Causal Inference}
The purpose of causal inference is to estimate the outcome shift (or effect) of different treatments. Let symbol $A$ denote a treatment that refers to an action that applies to a unit. For example, if we have a medicine $A$, let $A = 1$ denote applying medicine $A$ and $A = 0$ denotes not applying medicine $A$,  then $A = 1$ is a treatment, and the recovery of the patient is the outcome of the treatment $A = 1$. Under this condition, the aim of causal inference is to uncover the effect of applying treatment $A$. A counterfactual outcome is the potential outcome of an action that has not been taken. For example, if we take treatment $A = 1$, then the outcome of $A = 0$ is counterfactual. Then the average treatment effect (ATE) of treatment $A = 1$ could be written as:
\begin{equation}
    \textrm{ATE} = \mathbb{E}[Y(A = 1) - Y(A = 0)]
\end{equation}
where $Y(A = a)$ denotes the potential outcome of treatment $A = a$. If we have taken treatment $A = 1$, then $Y(A = 0)$ is the counterfactual outcome.

The goal of causal inference is to estimate the treatment effects given the observational data, which is usually incomplete in real-world scenarios due to the cost and moral problems. From a counterfactual perspective, we cannot always obtain a no-treatment outcome if we apply the treatment. Thus, we need to adopt causal inference to analyze the effect of a certain treatment.

\subsection{Causal Intervention}

 Causal intervention for machine learning aims to capture the causal effects of interventions (i.e., variables), and take advantage of causal relations in datasets to improve model performance and generalization ability. The basic idea of causal intervention is to use an adjustment strategy that modifies the graphical model and manipulates conditional probabilities to discover causal relationships among variables. In this section, we review two adjustment strategies: back-door adjustment and front-door adjustment.

\subsubsection{Back-door Adjustment}
Assume that we want to gauge the causal effect between $X$ and $Y$ by Bayes’ rules; we can have:
\begin{equation}
    P(Y\vert X) = \sum_{z} P(Y\vert X,z) P(z\vert X)
\end{equation}
This conditional distribution could not represent the true causal effect of $X$ on $Y$, due to the existence of back-door path $X\leftarrow Z \rightarrow Y$. Variable $Z$ here is a confounder that not only affects pre-intervention $X$ but also the outcome $Y$, which would make the conditional distribution a collective effect of $X$ and $Z$, and thus leads to spurious correlation. To eliminate the spurious correlation introduced by the back-door path, the back-door adjustment uses do-operator to calculate the intervened probability $P(Y\vert do(X))$ instead of the conditional probability $P(Y \vert X)$:
\begin{equation}
    P(Y\vert do(X)) = \sum_{z} P(Y \vert X,z) P(z)
    \label{backdoor adjust}
\end{equation}

\begin{figure}
    \begin{center}
        \includegraphics[width=7.5cm]{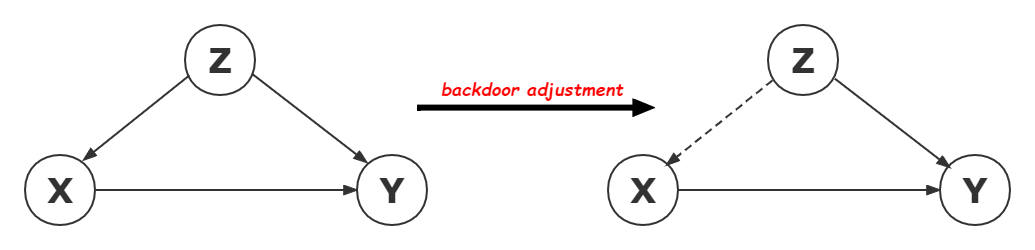}
    \end{center}
    \caption{An example of back-door adjustment, the back-door path from $X$ to $Y$ is blocked by cutting off the edge from $Z$ to $X$.}
    \label{fig:backdoor}
\end{figure}

Compare with $P(Y\vert X)$, $P(Y\vert do(X))$ replace the conditional distribution $P(z\vert X)$ with the marginal distribution $P(z)$. Fig. \ref{fig:backdoor} is a graphical view of the do-operator. The edge from $Z$ to $X$ is deleted in the intervened causal graph to block the back-door path  $X\leftarrow Z \rightarrow Y$; thus, $X$ and $Z$ become independent after the intervention. After the back-door adjustment, the intervened distribution $P(Y\vert do(X))$ can remove the spurious correlation between $X$ and $Z$ and calculate the true causal effect of variable $X$. Backdoor adjustment measures the causal effect of a variable by finding and blocking back-door paths points to it.

\subsubsection{Front-door Adjustment}

\begin{figure}[t]
    \centering
    \includegraphics[width=4cm]{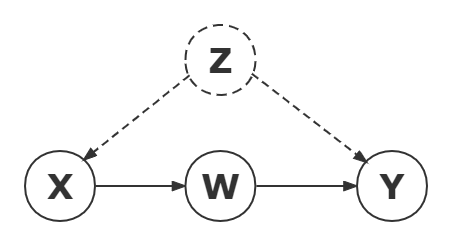}
    \caption{The back-door criterion is not satisfied since $Z$ is an unobserved variable.}
    \label{fig:frontdoor}
\end{figure}

The back-door criterion may not be satisfied in some causal graphical patterns (e.g., no back-door paths exist in causal graphs, or variables that block the back-door paths are unobserved). In such a case, the front-door adjustment pattern can be applied to estimate causal effects. As Fig. \ref{fig:frontdoor} shows, assuming that the variable $Z$ is an unobserved variable, the back-door adjustment becomes invalid because the marginal distribution $P(z)$ is not observed. However, if we have an observed mediator variable $W$ on the front-door path $X \rightarrow W \rightarrow Y$, then we can identify the effect of $X$ on $W$ directly since the back-door path from $X$ to $W$ is blocked by the collider at $Y$:
\begin{equation}
    P(W\vert do(X)) = P(W\vert X)
    \label{dox}
\end{equation}

Note that there is a back-door path from $W$ to $Y$ : $W \leftarrow X \leftarrow Z \rightarrow Y$, which can be blocked by applying back-door adjustment on $X$:
\begin{equation}
    P(Y \vert do(W)) = \sum_{x}P(Y \vert W,x)P(x)
    \label{dow}
\end{equation}

And the total effect of $X$ on $Y$ could be written by summing on $W$ :
\begin{small}
\begin{equation}
    P(Y \vert do(X)) =\\
    \sum_{w} P(Y \vert do(W=w)) P(W=w \vert do(X))
    \label{front-door adjustment}
\end{equation}
\end{small}

Then, the front-door adjustment formulation is obtained by applying Eq. \ref{dox}, Eq. \ref{dow}, and Eq. \ref{front-door adjustment}:
\begin{scriptsize}
\begin{small}
\begin{equation}
\begin{aligned}
P(Y\vert do(X))=\sum_{w}\sum_{x}P(Y\vert W=w,x)P(W=w\vert x)P(x)
\end{aligned}
\end{equation}
\end{small}
\end{scriptsize}
The front-door adjustment identifies the effect of $X$ on $Y$ by applying the do-operator twice, one at the mediator variable $W$ and the other at variable $X$ that blocks the back-door path. In this way, the unobserved variable $Z$ can be bypassed in intervention.

\subsubsection{Back-door or Front-door?}
The back-door adjustment requires us to determine what the confounder is in advance. Thus, the back-door adjustment is effective when the confounder is observable. However, in visual domains, data biases are complex, and it is hard to know and disentangle different types of confounders. Especially for some challenging tasks like the visual-linguistic question reasoning where the confounders in visual and linguistic modalities are not always observable. Therefore, the front-door causal intervention gives a feasible way to calculate $P(Y\vert do(X))$ when we cannot explicitly represent the confounder.

\section{Causality-aware Feature Learning}\label{sec4}
Traditional feature learning methods usually learn the spurious correlation introduced by confounders. This will reduce the robustness of models and make models hard to generalize across domains.
Causal reasoning, a learning paradigm that reveals the real causality from the outcome, overcomes the essential defect of correlation learning and learns robust, reusable, and reliable features. In this chapter, we review the recent representative causal reasoning methods for general feature learning, which mainly consist of three main paradigms: (1) Structural causal model (SCM) embedded, (2) Applying causal intervention/counterfactual, and (3) Markov boundary (MB) based feature selection.

For embedding the structural causal model (SCM), Mitrovic et al. \cite{mitrovic2020representation} proposed representation learning via invariant causal mechanisms (RELIC) to address self-supervised learning problems and achieved competitive performance in terms of robustness and out-of-distribution generalization on ImageNet. \citet{shen2020disentangled} proposed a disentangled generative causal representation (DEAR) learning method for causal controllable generation on both synthesized and real datasets.

To apply causal intervention or counterfactual inference for feature learning, \citet{huang2021deconfounded} proposed a causal intervention-based deconfounded visual grounding method to eliminate the confounding bias. \citet{zhang2020causal} present a causal inference based weakly-supervised semantic segmentation framework. \citet{tang2020long} present a causal inference framework that disentangles the paradoxical effects of the momentum to remove the confounder of long-tail classification. \citet{chen2020counterfactual} proposed a Counterfactual critic Multi-Agent Training (CMAT) approach to learn the visual context properly.

Causal feature selection aims to find a subset of features from a large number of predictive features to reduce computational cost and build predictive models for variables of interest.
Recent causality-based feature selection methods use Bayesian network (BN) and Markov boundary (MB) to identify potential causal features. BN is used as a DAG representing the causal relations between variables, and MB implies the local causal relationships between the class variable and the features in its MB. Since the BN of variables may be very large and hard to compute, current causal-based methods focus on identifying the MB as a variable or a subset of the MB. For example, \citet{wu2019accurate} introduced the PCMasking concept to explain a type of incorrect CI tests in MB discovery and proposed a CCMB algorithm to solve the incorrect test problem. \citet{yu2019learning} presented theoretical analyses on the conditions for MB discovery in multiple interventional datasets and designed an algorithm for learning MBs from multiple interventional datasets. \citet{yu2019multi} formulated the causal feature selection problem with multiple datasets as a search problem and gave the upper and lower bounds of the invariant set, then proposed a multi-source feature selection algorithm. \citet{yang2021towards} proposed the concept of N-structures and then designed an MB discovering subroutine to integrate MB learning with N-structures to discover MB while distinguishing direct causes from direct effects. \citet{yu2021multilabel} proposed a multi-label feature selection algorithm, M2LC, which learns the causal mechanism behind the data and is able to select causally informative features and visualize common features. \citet{guo2022error} proposed an error-aware Markov blanket learning algorithm to solve the conditional independence test error in causal feature selection. \citet{ling2020using} proposed an efficient local causal structure learning algorithm, LCS-FS, which speeds up parent and children discovery by employing feature selection without searching for conditioning sets. \citet{yu2022causal} proposed a multiple imputation MB framework MimMB for causal feature selection with missing data. MimMB integrates Data Imputation with MB Learning in a unified framework to enable the two key components to engage with each other.

Finding causal features improves the explanatory capability and robustness of models. Causal feature selection methods can provide a more convincing explanation for prediction than correlation-based methods. As the ICM principle implies, the underlying mechanism of the class variable can be learned from causal relations and thus can be transferred across different settings or environments. Although the existing causal feature learning methods achieve promising performance, most of them focus on general feature learning without considering a more specific problem, visual representation learning.

\section{Visual Representation Learning: State-of-the-art}\label{sec8}
Visual representation learning has made great progress in recent years, which can utilize spatial or/and temporal information to complete specific tasks, including visual understanding (object detection, scene graph generation, visual grounding, visual commonsense reasoning), action detection and recognition, and visual question answering, etc. In this section, we introduce these representative visual learning tasks and discuss the existing challenges and necessity of applying causal reasoning to visual representation learning.

\textbf{Object detection} aims to determine where objects are located in a given image (object localization) and to which category each object belongs to (object classification) and label them with rectangular bounding boxes (BBs) to show their confidence in existence. In image object detection, deep learning frameworks for object detection are divided into two types. The first type is to follow the traditional object detection process, generating region proposals firstly and then classifying each proposal into a different object class. The other type is to treat object detection as a regression or classification problem and adopt a unified framework to directly obtain the final predictions (category and location). Region proposal-based methods mainly include R-CNN \cite{girshick2014rich}, Spatial Pyramid Pooling (SPP-net) \cite{he2015spatial}, Fast R-CNN \cite{girshick2015fast}, Faster R-CNN \cite{ren2015faster}, Feature Pyramid Network (FPN) \cite{lin2017feature}, Region-based Fully Convolutional Network (R-FCN) \cite{dai2016r}, and Mask R-CNN \cite{he2017mask}, some of which are interrelated (e.g., SPP-net modifies R-CNN with an SPP layer). Based on regression/classification, the methods mainly include MultiBox \cite{erhan2014scalable}, AttentionNet \cite{yoo2015attentionnet}, G-CNN \cite{najibi2016g}, YOLO \cite{redmon2016you}, Single Shot MultiBox Detector (SSD) \cite{liu2016ssd}, YOLOv2 \cite{redmon2017yolo9000}, Deeply Supervised Object Detector (DSOD) \cite{shen2017dsod} and Deconvolution Single Shot Detector (DSSD) \cite{fu2017dssd}. The correlations between these two pipelines are connected by anchors introduced in Faster R-CNN. In video saliency object detection, extending state-of-the-art saliency detectors from images to videos is challenging. \citet{li2018flow} presented a flow-guided recurrent neural encoder (FGRNE), which works by enhancing the temporal coherence of the per-frame feature by exploiting both motion information in terms of optical flow and sequential feature evolution encoding in terms of LSTM networks. \citet{li2019motion} developed a multi-task motion-guided video salient object detection network, which learns to accomplish two sub-tasks using two sub-networks, one sub-network is for salient object detection in still images and the other one is for motion saliency detection in optical flow images. \citet{yan2019semi} presented an effective video saliency detector that consists of a spatial refinement network and a spatiotemporal module. By utilizing the generated pseudo-labels together with a part of manual annotations, the detector can learn spatial and temporal cues for both contrast inference and coherence enhancement. For video salient object detection, how to effectively take object motion into consideration and obtain robust spatial-temporal information is crucial in video salient object detection. However, some non-object, occlusion, motion blur, and lens movement make the model hard to concentrate on the true interesting object area.

\begin{figure}[t]
    \centering
    \includegraphics[width=6cm]{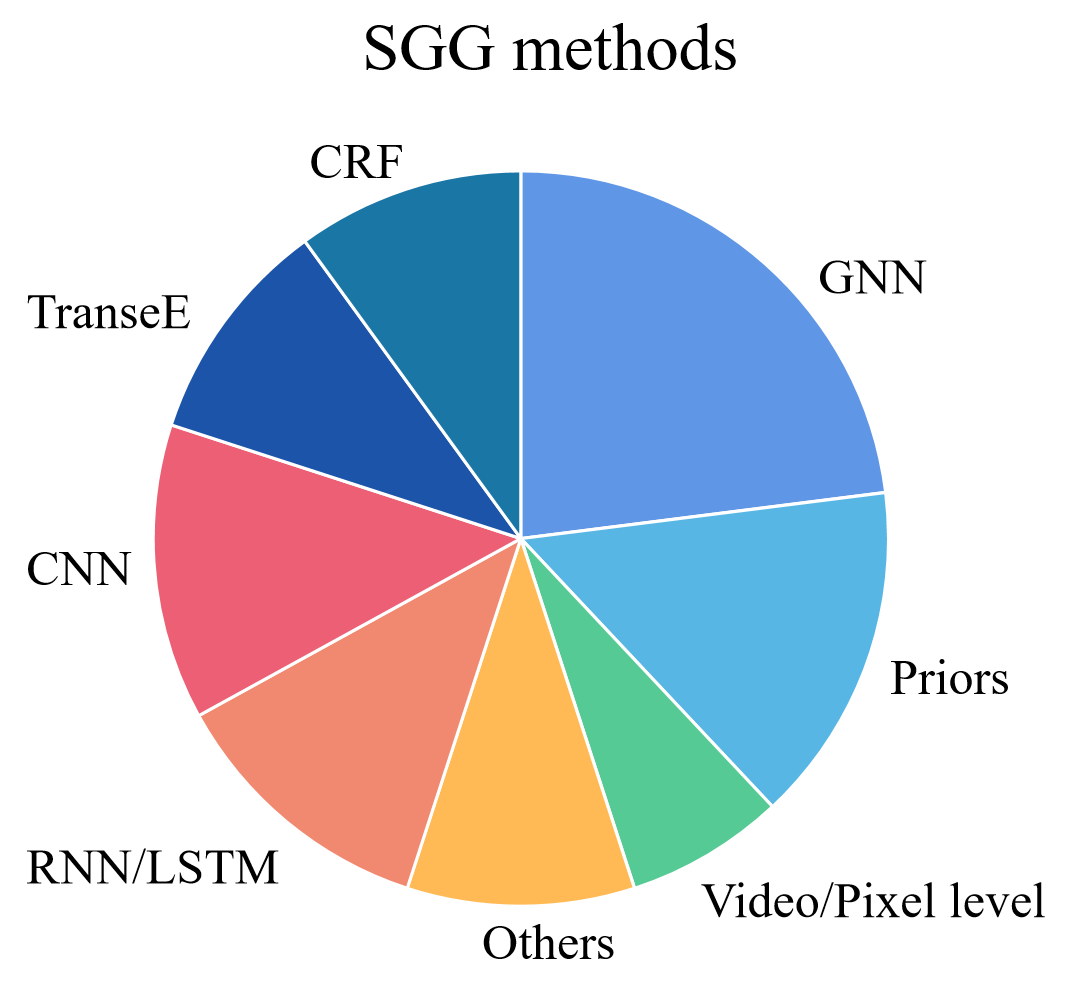}
    \vspace{1.0em}
    \caption{The classification and statistics of scene graph generation (SGG) methods.}
    \label{fig:SGG methods}
\end{figure}

\textbf{Scene graph generation (SGG)} aims to describe object instances and relations between objects in a scene. With its powerful representation ability, SGG can encode images \cite{armeni20193d,johnson2015image} and videos \cite{wang2020storytelling,qi2018scene} as its abstract semantic elements without any restrictions on the attributes, types, and relations between objects. Therefore, the task of SGG is to build a graph structure that associates its nodes and edges well with objects in the scene and their relations, where the key challenge task is to detect/recognize relations between objects. Currently, SGG can be divided into two classes: (1) With facts alone and (2) Introducing prior information. Besides, these SGG methods pay more attention to the methods with facts alone, including CRF-based (conditional random field) SGG \cite{johnson2015image, dai2017detecting}, VTransE-based (visual translation embedding) SGG \cite{zhang2017visual, hung2019union}, RNN/LSTM-based SGG \cite{chen2019panet, tang2019learning}, Faster RCNN-based SGG \cite{li2017vip,liang2019vrr}, GNN \cite{li2018factorizable, qi2019attentive}, etc. Furthermore, SGG adds different types of prior information, such as language priors \cite{lu2016visual}, knowledge priors \cite{chen2019knowledge, gu2019scene}, visual contextual information \cite{zellers2018neural}, visual cue \cite{plummer2017phrase}, etc. Fig. \ref{fig:SGG methods} shows the related work on SGG, and it can be clearly seen that most of the methods use the GNN model or introduce relevant prior information when conducting SGG. Existing SGG methods are still far from building a practical knowledge base. There exists a serious conditional distribution bias of the relationship in SGG methods. For example, knowing that the subject and object are \emph{person} and \emph{head}, it is easy to guess that the relationship is that a \emph{person has a head}.

\textbf{Visual grounding} usually involves two modalities, visual and linguistic data. This task aims to locate the target object in the image according to the corresponding object description (title or description) and the given image. When locating the target object, it is necessary to understand the input description information, and integrate the information of the visual modality for localization prediction. Currently, visual grounding methods can be classified into three types: fully supervised \cite{yang2019cross,lin2021scene,liu2021refer,sun2021iterative,kamath2021mdetr, deng2021transvg, wu2020tree, chen2021ref}, weakly supervised \cite{ wu2020reinforcement}, and unsupervised \cite{yeh2018unsupervised}. First, the fully supervised methods contain box annotations with object-phrase information. This method can be further divided into two-stage methods \cite{lin2021scene,liu2021refer,kamath2021mdetr, chen2021ref} and one-stage method \cite{deng2021transvg}. The two-stage approach is to extract candidate proposals and their features in the first stage through a Region Proposal Network (RPN) \cite{girshick2015fast} or traditional algorithms (Edgebox \cite{zitnick2014edge}, Selective Search \cite{uijlings2013selective}). Second, weakly supervised methods \cite{liu2021relation,wang2021improving} only have images and corresponding sentences, and no box annotations for object-phrases in the sentences. Due to the lack of mapping between phrases and boxes, weak supervision will additionally design many loss functions, such as designing reconstruction loss, introducing external knowledge, and designing loss functions based on image-caption matching. Third, there is no image-sentence information in the unsupervised method. \citet{wang2019phrase} used off-the-shelf approaches to detect objects, scenes and colors in images and explore different approaches to measuring semantic similarity between the categories of detected visual elements and words in phrases. To locate the object instance described by a natural language referring expression in an image, some referring expression comprehension methods are proposed. \citet{yang2019dynamic} proposed a dynamic graph attention network to perform multi-step reasoning by modeling the relationships among the objects in the image and the linguistic structure of the expression. \citet{yang2019cross} proposed a Cross-Modal Relationship Extractor (CMRE) to adaptively highlight objects and relationships with a cross-modal attention mechanism, and represented the extracted information as a language-guided visual relation graph. Furthermore, \citet{yang2020relationship} proposed a cross-modal relationship extractor to adaptively highlight objects and relationships (spatial and semantic relations) related to the given expression with a cross-modal attention mechanism, and represent the extracted information as a language-guided visual relation graph. \citet{yang2020graph} proposed a scene graph-guided modular network (SGMN), which performed reasoning over a semantic graph and a scene graph with neural modules under the guidance of the linguistic structure of the expression. However, due to the existence of linguistic and visual biases, most visual grounding models are heavily dependent on specific datasets, without good transfer ability and generalization performance.

Due to the success of BERT-related models in the field of NLP, researchers have begun to focus on more a challenging multi-modal reasoning task, \textbf{Visual Commonsense Reasoning (VCR)}. The VCR task needs to combine image information with the understanding of questions, and obtain the correct answer as well as the reasoning process based on the commonsense. Given an image, the image contains a series of bounding boxes with labels. In general, VCR can be divided into two sub-tasks: ${Q\rightarrow A}$ task is choosing an answer based on the question; and ${QA\rightarrow R}$ task is reasoning based on the question and the answer, explaining why the answer was chosen. Due to the challenging nature of VCR, there are actually relatively few existing studies. Some of them resort to designing specific model architectures \cite{zellers2019recognition,wu2019connective,yu2019heterogeneous,lin2019tab,zhang2021multi}. R2C \cite{zellers2019recognition} implemented this task with a three-step approach, associating text with objects involved, linking answers with corresponding questions and objects, and finally, reasoning about shared representations. Inspired by brain neuron connectivity, CCN \cite{wu2019connective} dynamically modeled the visual neuron connectivity, which is contextualized by the queries and responses. HGL \cite{yu2019heterogeneous} leveraged visual answering and dual question answering heterogeneous graphs to seamlessly connect vision and language. \citet{zhang2021multi} proposed a multi-level counterfactual contrastive learning network for VCR by jointly modeling the hierarchical visual contents and the inter-modality relationships between the visual and linguistic domains. Recently, BERT-based pre-training methods have been extensively explored in vision and language domains. In general, most of them adopt a pre-training-then-transfer scheme and achieve significant performance improvements on the VCR \cite{lu2019vilbert,chen2020uniter,su2019vl} benchmarks. These models are usually pre-trained on large-scale multi-modal datasets (e.g., concept captioning \cite{sharma2018conceptual}) and then fine-tuned on VCR. At present, the promising performance of VCR is generally attributed to the pre-trained big model and the prior external knowledge. Compared with simple vision-linguistic domain tasks, the introduction of external knowledge brings new challenges: (1) How to retrieve limited supporting knowledge from external knowledge bases that contain massive data. (2) How to effectively integrate external knowledge with visual and linguistic features. (3) The reasoning process gives interpretability needs supporting facts, which depends heavily on the language structure design.

The task of \textbf{action detection and recognition} includes two aspects, one is to identify all action instances in the video, and the other one is to localize actions spatially and temporally. Nowadays, spatial-temporal action detection or recognition models can be divided into two categories, the first one \cite{carreira2017quo,sun2018actor,lin2019tsm,wu2019long,yang2020temporal,feichtenhofer2020x3d,feichtenhofer2019slowfast,feichtenhofer2020x3d,bao2021evidential,liu2021temporal_3,aich2021spatio} is to model spatial-temporal relationships based on Convolutional Neural Networks (CNNs), and the other one \cite{tan2021relaxed,bertasius2021spacetime,wang2021oadtr,zhang2021temporal,bertasius2021space} is based on video transformer structures. Besides, the skeleton-based models \cite{yan2018spatial,yan2018spatial,si2019attention,shi2019two,lin2021end} have recently attracted great attention. \citet{sun2018actor} proposed an actor-centric relational network (ACRN), which used two-stream to extract the central character feature and global background information from the input clip, and then performed feature fusion for action classification. \citet{feichtenhofer2019slowfast}  proposed a two-stream model named SlowFast networks that contains a Slow pathway and a Fast pathway. \citet{bertasius2021space} simply extended the ViT \cite{dosovitskiy2020image} design to video by proposing several scalable schemes for space-time self-attention. \citet{arnab2021vivit} proposed pure-transformer architectures for video classification, including several variants of the model by factorizing the spatial and temporal dimensions of the input video. Although great progress has been made in spatial-temporal action detection and recognition based on the CNNs or transformer models, there exist some critical problems in terms of the robustness and the transferability of the models. The existing action detection and recognition models rely heavily on scenes and objects. When a model is well-trained in one dataset, it is hard to generalized to another dataset with different scenes. Additionally, the methods are easily focused on some static appearance or background information rather than the true motion area due to the essential correlation learning in most of the models. This may be harmful to the reliability of the model, as well as the robustness of the learned spatial-temporal representations. Causal reasoning has the powerful ability to uncover the underlying structural knowledge about human actions that build a strong cognitive model that can fully discover causality and spatial-temporal relations.

\textbf{Visual question answering (VQA)} is a vision-language task that has received much attention recently. The objective of VQA is: given the image/video and a related question, a machine needs to reason over visual elements and general knowledge to infer the correct answer. The attention mechanism is widely used in VQA models, which aim to focus on the critical part of the image and question, and find cross-modality correlations. The UpDn \cite{anderson2018bottom} framework is a typical conventional VQA method based on attention, which uses a top-down attention LSTM \cite{hochreiter1997long} for the fusion of visual and linguistic features. Besides using LSTM, the transformer \cite{vaswani2017attention} can also be adapted to the VQA task, thanks to its powerful scaled dot-product attention block. VLP (Visual-Language Pre-training) models based on BERT \cite{devlin2018bert} show remarkable performance in the VQA task. ViLBERT \cite{lu2019vilbert} is a BERT-based visual and language pre-training framework, which uses a self-attention transformer block \cite{vaswani2017attention} to model in-modality relation and develop a co-attention transformer block to compute cross-modality attention score, and it finally achieves a state-of-the-art on four visual-language tasks including VQA at that time. Compared with the Image QA \cite{antol2015vqa,yang2016stacked,anderson2018bottom,cao2021knowledge,cao2021linguistically}, the video question answering (VideoQA) task is much more challenging due to the existence of extra temporal information. To accomplish the VideoQA problem, the model needs to capture spatial, temporal, visual, and linguistic relations to reason about the answer. To explore relational reasoning in VideoQA, Xu et al. \cite{xu2017video} proposed an attention mechanism to exploit the appearance and motion knowledge with the question as guidance. Later on, some hierarchical attention and co-attention-based methods are proposed to learn appearance-motion and question-related multi-modal interactions. Le et al. \cite{le2020hierarchical} proposed a hierarchical conditional relation network (HCRN) to construct sophisticated structures for representation and reasoning over videos. Jiang et al. \cite{jiang2020reasoning} introduced a heterogeneous graph alignment (HGA) network. Huang et al. \cite{huang2020location} proposed a location-aware graph convolutional network to reason over detected objects. Lei et al. \cite{lei2021less} employed sparse sampling to build a transformer-based model named CLIPBERT and achieve end-to-end video-and-language understanding. Liu et al. \cite{liu2021hair} proposed a hierarchical visual-semantic relational reasoning (HAIR) framework to perform hierarchical relational reasoning. Although hierarchical attention mechanisms successfully improve the visual-language task performance, these models remain with a strong reliance on modality bias \cite{goyal2017making, agrawal2018don} and tend to capture the spurious linguistic or visual correlations within the images/videos, and thus fail to learn the multi-modal knowledge with good generalization ability and interpretability.

\begin{table*}[!htb]
\renewcommand\arraystretch{1}\renewcommand\tabcolsep{1pt}
    \centering
        \caption{Recent causal visual representation learning methods, tasks, basic models, and causal reasoning types.}
    \label{tab:causalvqa}
    \resizebox{\textwidth}{40mm}{
    \begin{tabular}{@{}lccc@{}}
\toprule
Author \& Year &  \multicolumn{1}{c}{Tasks}  & \multicolumn{1}{c}{Basic Models}   & \multicolumn{1}{c}{Types of Causal Reasoning} \\ \midrule
\citet{agarwal2020towards} 2020&   Visual Question Answering &           GAN \cite{goodfellow2014generative} &                               Counterfactual Sample Synthesising                  \\
\citet{chen2020counterfactual} 2020 &  Visual Question Answering & CCS \cite{chen2020counterfactual} & Counterfactual Sample Synthesising
\\
\citet{zhang2020causal} 2020 & Weakly-Supervised Semantic Segmentation & Pseudo-Mask Generation & Backdoor Adjustment
\\
\citet{tang2020unbiased} 2020 & Scene Graph Generation &  Unbiased Training & Causal Inference
\\
\citet{tang2020long} 2020 & Long-Tailed Classification & De-confounded Training & Causal Inference
\\
\citet{wang2020visual} 2020 & IC\& VQA \& VCR & R-CNN & Causal Intervention
\\
\citet{yue2020interventional} 2020 & Few-Shot Learning & - & Backdoor Adjustment
\\
\citet{hu2021distilling} 2021 & Class Incremental Learning & - & Causal Inference
\\
\citet{yue2021transporting} 2021 & Unsupervised Domain Adaptation & GAN \& VAE & Causal Inference
\\
\citet{mao2021generative} 2021 & Domain Adaptation & GAN & Causal Inference
\\
\citet{tang2021adversarial} 2021 & Adversarial Defense & Instrumental Variable Intervention & Causal Intervention
\\
\citet{wang2021causal} 2021 &Visual Recognition & Adversarial Training & Causal Inference
\\
\citet{yue2021counterfactual} 2021 & Zero-shot \& Open-set Visual Recognition & - & Counterfactual Inference
\\
\citet{huang2021deconfounded} 2021 & Visual Grounding & CNN, BERT \cite{devlin2018bert}, Attention \cite{vaswani2017attention} & Causal Inference
\\
\citet{zhang2021devlbert} 2021 &   Visual Question Answering &    BERT \cite{devlin2018bert}                       &   Backdoor Adjustment                                    \\
\citet{yang2021causal} 2021 &      Visual Question Answering &      Attention \cite{vaswani2017attention}                 &       Front-door Adjustment                               \\

\citet{niu2021counterfactual} 2021& Visual Question Answering & CF-VQA \cite{niu2021counterfactual} & Counterfactual Inference \\

\citet{li2022invariant} 2022& Video Question Answering & IGV \cite{li2022invariant} & Invariant Grounding \\

\citet{liu2022contextual} 2022& Visual Recognition & CNN/Transformer, CCD \cite{liu2022contextual} & Causal Context Debiasing \\

\citet{liu2022towards} 2022& Motion Forecasting & Encoder-Decoder \cite{liu2022towards} & Causal Invariant Learning \\

\citet{lv2022causality} 2022& Domain Generalization & CIRL \cite{lv2022causality} & Causal Intervention \\

\citet{LinCLY22} 2022& Video Anomaly Detection & UVAD \cite{LinCLY22} & Causal Intervention, Counterfactual Inference \\

\citet{LinWCLY22} 2022& Salient Object Detection & USOD \cite{LinWCLY22} & Causal Intervention \\

\citet{liu2022cross} 2022& Event-level Video Question Answering & Transformer, CMCIR \cite{liu2022cross} & Causal Intervention \\
 \bottomrule
\end{tabular}
}
\end{table*}

\section{Causality-aware Visual Representation Learning}
According to the above-discussed visual representation learning methods, the current machine learning, especially representation learning, faces several challenges: (1) Lack of interpretability, (2) Poor generalization ability, and (3) Over-reliance on correlations of data distribution. Causal reasoning offers a promising alternative to address these challenges. The discovery of causality helps to uncover the  causal mechanism behind the data, allowing the machine to understand better why and to make decisions through intervention or counterfactual reasoning. Since Section 3 has reviewed the recent causal reasoning methods for general feature learning, it provides a good theoretical basis for further research on causal reasoning with specific visual representation learning tasks. In this section, we summarize some recent approaches for causal visual representation learning, as shown in Table \ref{tab:causalvqa}. The causal visual representation learning is an emerging research topic and has appeared since the 2020s. The related tasks can be roughly categorized into several main aspects: (1) Causal visual understanding, (2) Causal visual robustness, and (3) Causal visual question answering. In this section, we discuss these three representative causal visual representation learning tasks.

\subsection{Causal Visual Understanding}
Visual understanding contains several tasks, such as object detection, scene graph generation, visual grounding, visual commonsense reasoning, etc. However, some challenges exist in these tasks: (1) For image/video object saliency detection, some non-object, occlusion, motion blur, and lens movement make the model hard to concentrate on the true interesting object area. To this end, causal reasoning can make the model focus on the essential interesting object area by learning robust and reliable visual representations. (2) For the scene graph generation (SGG) problem that contains superficial bias and insufficient generalization ability, causal reasoning can be introduced to mitigate these problems well. For example, an item such as \emph{towel} is used to bathe in the bathroom, but is used to wash the face in the office. Introducing causal reasoning into SGG can generalize the functionality of an item to different scenarios. (3) Due to the existence of linguistic and visual biases, most visual grounding models are heavily dependent on specific datasets without good transfer ability and generalization performance. This problem can be mitigated by causal reasoning methods, which learn robust and transferable features to mitigate
the visual and linguistic biases.  (4) For visual commonsense reasoning (VCR), linguistic biases may directly affect the reasoning performance. Generally, the superficial correlations captured by the existing VCR models can be mitigated by introducing causality that integrates external knowledge and visual and linguistic features into a robust and discriminative representation space. Non-causal visual understanding methods are easily affected by confounders in visual content. Illumination, position, backgrounds, co-occurrence of objects, and other visual factors are confounders that are inevitable in common settings. With traditional correlation learning, spurious correlations introduced by the confounders degrade the robustness of representation learning. For example, since the co-occurrence of ``bird" and ``sky" are high, the model would learn a strong correlation between them. Thus, when seeing a picture of a floating balloon that also contains ``sky", it would also make a confident prediction that it is a picture of a bird.

Causal reasoning provides a good solution to address the above problem. By replacing the conditional distribution with the intervened distribution, the spurious correlation can be eliminated, and the machine can learn the real causality. Applying intervention in the training procedure is a widely used implementation of the causal intervention. In the visual recognition task, \citet{wang2021causal} combined adversarial training with causal intervention, modeled the different causal effects of mediators and confounders, and designed an adversarial training pipeline to improve the effect of mediators while suppressing the effect of confounders. \citet{yue2021counterfactual} applied counterfactual inference to zero-shot and open-set visual recognition by proposing a generative causal model to generate counterfactual samples. A confounding effect also exists in the visual grounding task, \citet{huang2021deconfounded} proposed a deconfounded visual grounding framework by conducting interventions on linguistic features. For the weakly-supervised semantic segmentation task,  \citet{zhang2020causal} used the structural causal model to formulate the causalities between components, then constructed a confounder set and removed confounders by back-door adjustment. The prior bias is also a non-trivial problem in the SSG (scene graph generation) task. To reduce the negative impact of training bias in scene graph generation, \citet{tang2020unbiased} built a causal graph and extracted counterfactual causality from the trained graph to infer the causal effects of training bias and then remove the negative bias. Besides, causal reasoning can replace the traditional re-weighting and re-sampling methods in resolving long-tailed distribution problems. \citet{tang2020long} analyzed that the momentum in SGD introduces the unbalanced sample distribution and then proposed to use counterfactual inference in the test stage to detect and remove the causal effect of the momentum item. \citet{wang2020visual} proposed an unsupervised commonsense learning framework to learn intervened visual features by back-door adjustment, which can be used in the downstream task as image captioning, visual question answering, and visual commonsense reasoning. \citet{liu2022contextual} established a Structural Causal Model (SCM) to uncover the causal relevance among contextual priori, object feature, contextual bias, and final prediction in multi-target visual tasks. \citet{liu2022towards} introduced a causal formalism of motion forecasting, which casts the problem as a dynamic process with three groups of latent variables, namely invariant variables, style confounders, and spurious features. \citet{LinCLY22} proposed a causal graph to analyze the confounding effect of the pseudo label generation process for unsupervised video anomaly detection. \citet{LinWCLY22} proposed a causal-based debiasing framework to disentangle the unsupervised salient object detection from the impact of contrast distribution bias and spatial distribution bias.

\subsection{Causal Visual Robustness}
The ubiquitous spurious correlation learned by deep learning models reduces the model robustness, which is a potential vulnerability of the conventional deep learning paradigm. In this perspective, the causal learning paradigm can be introduced to avoid the presence of confounding effects and make the model more robust \cite{ren2022towards}.

Confounders are widespread in visual robustness problems, including few-shot learning, class-incremental learning, domain adaptation, generative model, etc. \citet{yue2020interventional} uncovered that pre-trained knowledge is a confounder in few-shot learning and developed a few-shot learning paradigm by introducing back-door adjustment to control the pre-trained knowledge. The confounding effect can be leveraged by attackers,  \citet{tang2021adversarial} proposed an instrumental variable \cite{bowden1990instrumental} estimation-based causal regularization method for adversarial defense. \citet{hu2021distilling} explained the catastrophic forgetting effect in class-incremental learning in terms of causality: the causal effect of old data is zero, and then proposed distilling the causal effect of old data by controlling the collider effect of the causal graph. As ICM inferred, causal mechanisms could be invariant across domains; hence, learning invariant causal knowledge is likely to be superior in robustness. To learn cross-domain knowledge, \citet{yue2021transporting} disentangled semantic attributes in images into causal factors and used CycleGAN \cite{zhu2017unpaired} to generate counterfactual samples in the counterpart domain, then exploited the counterfactual sample and a latent variable encoded by VAE \cite{kingma2013auto} as proxy variables of an unobserved attribute for intervention. Apart from generating counterfactual samples, intervention can also be implemented by the generative method. \citet{mao2021generative} argued that conventional randomized control trials and intervention approaches could hardly be used in naturally collected images, then introduced a framework performing interventions on realistic images by steering generative models to generate intervened distribution. \citet{lv2022causality} introduced a Causality Inspired Representation
Learning (CIRL) algorithm that enforced the representations to satisfy three properties, and then used them to simulate causal factors.

\subsection{Causal Visual Question Answering}
For visual question answering, the real causality behind the visual-linguistic modalities and the interaction between the
appearance-motion and language knowledge are neglected in most of the existing methods. In recent works, the purpose of introducing causality into visual question-answering tasks is to reduce language bias in VQA tasks. Strong correlations between the question and the answer will make VQA models rely on spurious correlations without concerning visual knowledge. For example, since the answer to the question ``What is the color of the apple ?'' is ``red'' in most cases, the VQA model will easily learn the correlation between the word ``apple'' and the word ``red''. Thus, when given an image of a green apple, the model still predicts the answer ``red'' with strong confidence. Although simply balancing the dataset \cite{zhang2016yin, goyal2017making} can partly mitigate the linguistic bias, the spurious correlation still exists in the model. From this perspective, the causality-based solution is better than simply balancing the data, since the causal reasoning cuts off the superficial correlations and makes the VQA models focus on the real causality.

\begin{figure*}
    \centering
    \includegraphics[width=1\textwidth]{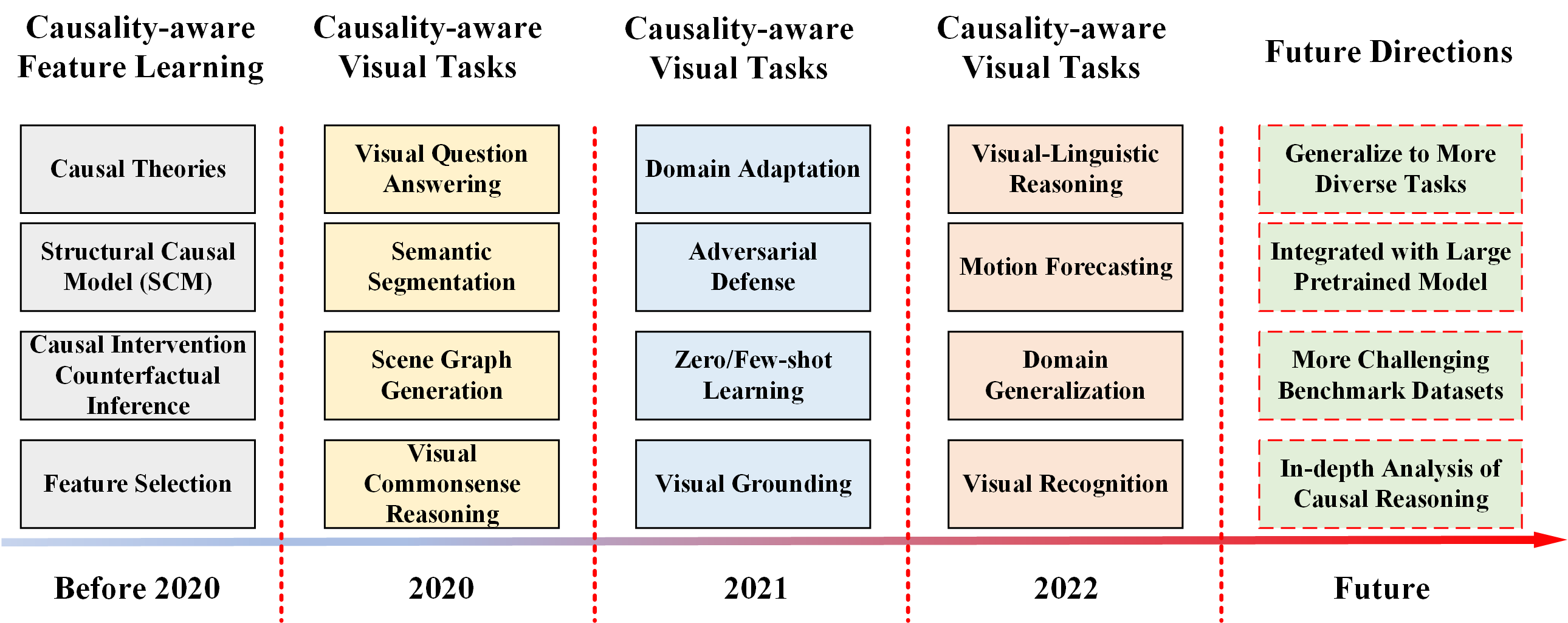}
    \caption{The development timeline of causal visual representation learning including the past, current, and future directions.}
    \label{fig:timeline}
\end{figure*}

Constructing a confounder set has been commonly used in causal intervention practice. VC-RCNN \cite{wang2020visual} constructed an object level visual confounder set for performing back-door adjustment in a visual task. Following VC-RCNN, DeVLBert \cite{zhang2021devlbert} treated nouns in linguistic modality as confounders and constructed language confounder sets using their average Bert representation vectors. Besides, DeVLBert incorporated the intervention into Bert's \cite{devlin2018bert}  pre-training process and combined mask modeling objective with causal intervention. As another implementation of the intervention, \citet{yang2021causal} designed the In-Sample attention and Cross-Sample attention module to conduct front-door adjustment, where the In-Sample attention module approximates probability $P(W = w\vert x)$, and the Cross-Sample attention module approximates probability $P(x)$. Using these attention modules, a cross modality causal attention network was proposed for the VQA task by combining causal attention with the previous LXMERT \cite{tan2019lxmert} framework. Counterfactual-based solutions are also worth noting. \citet{agarwal2020towards} proposed a counterfactual sample synthesizing method based on GAN \cite{goodfellow2014generative}. Overcoming the complexity of the GAN based synthesising method, \citet{chen2020counterfactual} tried to replace critical objects and critical words with mask tokens and reassigned an answer to the synthesis of counterfactual QA pairs. Apart from sample synthesizing methods, \citet{niu2021counterfactual} developed a counterfactual VQA framework that reduces multi-modality bias by using the total indirect effect (TIE) \cite{pearl2009causality} for final inference. By blocking the direct effect of one modality, the TIE measures the total causal effect of the question and visual information, thus reducing language bias in the VQA. Li et al. \cite{li2022invariant} proposed an Invariant Grounding for VideoQA (IGV) to force the VideoQA models to shield the answering process from the negative influence of spurious correlations, which significantly improves the reasoning ability. \citet{liu2022cross} proposed a causality-aware event-level visual question answering framework named Cross-Modal Causal Relational Reasoning (CMCIR) to discover true causal structures via causal intervention on the integration of visual and linguistic modalities.

\begin{table*}[!htb]

\renewcommand\arraystretch{1}\renewcommand\tabcolsep{10pt}
    \centering
    \caption{Current causal image question answering datasets. }
    \label{tab:vqadataset}
    \begin{scriptsize}
\begin{tabular}{@{}lcccc@{}}
\toprule
Datasets &  \multicolumn{1}{c}{Image Source}  & \multicolumn{1}{c}{Split (train/val/test)}   & Is Collected?& \multicolumn{1}{c}{Rebalanced?} \\ \midrule
VQA v1 \cite{Antol_2015_ICCV} & COCO \cite{chen2015microsoft} & 614K/-/- & Yes & No \\

VQA v2 \cite{goyal2017making} & COCO \cite{chen2015microsoft} & 443K/214K/453K & No & Yes \\

VQA-CP v1 \cite{agrawal2018don} & COCO \cite{chen2015microsoft} &
 245K/-/125K &  Yes & No \\

VQA-CP v2 \cite{agrawal2018don} & COCO \cite{chen2015microsoft} & 438K/-/220K  & No & Yes \\

IV-VQA \cite{agarwal2020towards} & COCO \cite{chen2015microsoft} & 257K/11.6K/108K & No & Yes \\

CV-VQA \cite{agarwal2020towards} & COCO \cite{chen2015microsoft} & 8.5K/0.4K/3.7K & No & Yes \\

AVQA \cite{Li_2021_ICCV} & Various &  142.1K/8.7K/26.4K & Yes & Yes \\

 \bottomrule
\end{tabular}
    \end{scriptsize}
\end{table*}

\subsection{Comparisons and discussions}
To summarize the development line and the current state of causal visual representation learning, we show the development situation of causal visual representation learning in Fig. \ref{fig:timeline}, including the past, current, and future directions. Although the above-mentioned causal visual representation learning methods successfully apply causal reasoning methods to uncover causal mechanisms and achieve promising results, causal reasoning for visual representation learning is still in its infancy stage with many challenges. Firstly, the existing causal visual representation tasks are limited to several computer vision tasks without being applied to more diverse and challenging tasks such as video understanding, human-computer interaction, urban computing, etc. It should be noticed that recent large pre-trained vision-language models like CLIP \cite{radford2021learning} have shown great potential in learning representations that are transferable across a wide range of downstream tasks. Different from traditional representation learning, which is based mostly on discretized labels, popular prompt learning adopts vision-language pre-training and aligns images/videos/texts in a common feature space, which allows zero-shot transfer to any downstream task via prompting. Therefore, how to apply causality-ware knowledge to prompt learning may be a potential direction. Secondly, causal reasoning has been burgeoned for many visual learning tasks. So far, the existing evaluation datasets are still as traditional datasets for correlation learning without proper large-scale benchmarking datasets and pipelines specified for causal reasoning. Thirdly, most of the existing methods focus on causality discovery on either visual or linguistic modality without considering both of them. Therefore, a more in-depth analysis of the relations between causal reasoning and visual representation learning is required.

\begin{table*}[!htb]
\renewcommand\arraystretch{1.5}
    \centering
    \caption{Video question answering (VideoQA) benchmarks related with reasoning. OC and OG denote open-ended question answering as problem of classification and generation respectively. MC stands for multi-choice QA.}
    \label{tab:videodatasets}
\resizebox{\textwidth}{15mm}{
    \begin{tabular}{@{}lcccccccc@{}}
\toprule
Datasets &  \multicolumn{1}{c}{Topic}  & \multicolumn{1}{c}{QA Pairs}   & \multicolumn{1}{c}{QA Task}& Annotation & Real World & \multicolumn{1}{c}{Diagnostic Annotations} & Counterfactual & Balanced \\ \midrule

CLEVRER \cite{yi2019clevrer} &  Object Collision & 282K &MC& Auto &  & \checkmark & \checkmark & \checkmark \\

VQuAD \cite{gupta2022vquad} & Object Movement & 1.35M &MC& Auto & & \checkmark & &  \checkmark \\

ComPhy \cite{chen2022comphy}& Hidden Physical Properties & 100K &OC\&MC& Auto & &\checkmark &\checkmark &\checkmark \\

AGQA \cite{grunde2021agqa} & Natural Video Scenes & 192M &OC\&Compositional & Auto & \checkmark & & & \\

SUTD-TrafficQA \cite{xu2021sutd} & Traffic Events & 62K &MC\&OC& Human & \checkmark & & \checkmark & \checkmark \\

NExT-QA \cite{xiao2021next} & Causal and Temporal Interactions & 52K &MC\&OG& Human & \checkmark & &\checkmark &\checkmark \\
 \bottomrule
\end{tabular}
}
\end{table*}

\begin{figure}
    \centering
    \includegraphics[scale=0.25]{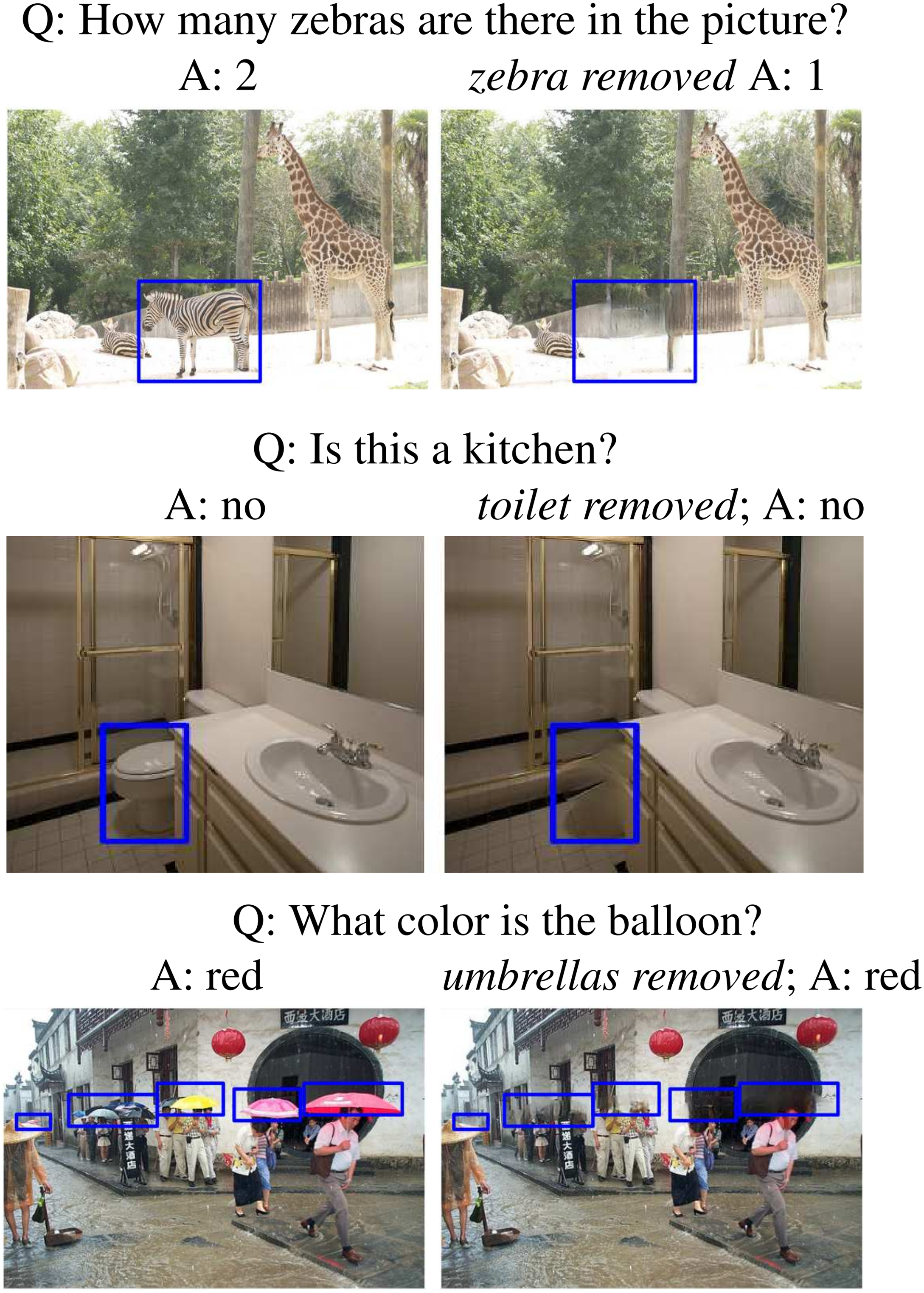}
    \caption{Examples in IV-VQA and CV-VQA datasets \cite{agarwal2020towards}.}
    \label{fig:IV-VQA}
\end{figure}

\section{Related Causal Datasets}\label{sec11}

Correlation-based models may perform well in existing datasets, not because these models have a strong reasoning capability, but because these datasets cannot fully support the evaluation of the models' reasoning capability. Spurious correlations in these datasets can be exploited by the model to cheat, which means that the model just concentrates on superficial correlation learning, not real causal reasoning, only approximating the distribution of the dataset. For example, in the VQA v1.0 \cite{Antol_2015_ICCV} dataset for the VQA task, the model simply answers ``yes'' when seeing the question ``Do you see a...'', which will achieve nearly 90\% accuracy. Due to this shortcoming in current datasets, researchers need to build benchmarks that can evaluate the true causal reasoning capability of models. In this section, we take image question answering benchmarks and video question answering benchmarks as examples to analyze the current research situation of related causal reasoning datasets and give some future directions.

\begin{figure}
    \centering
    \includegraphics[width=0.48\textwidth]{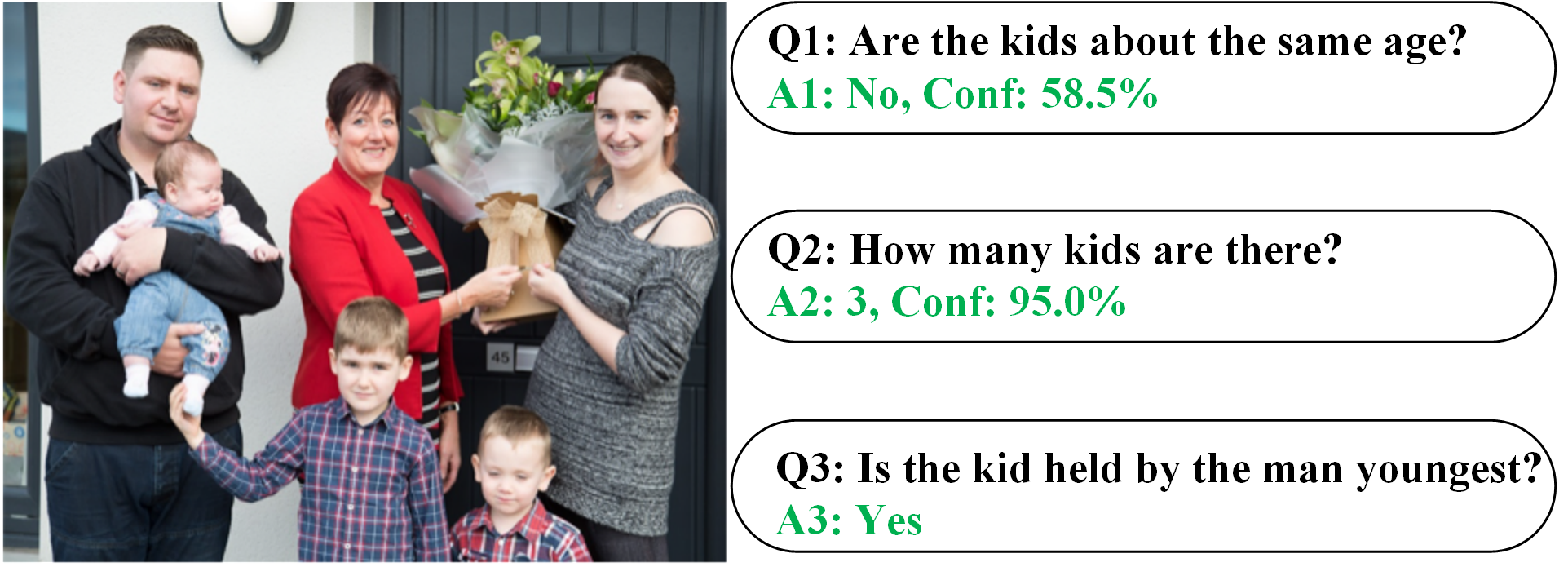}
    \caption{The example QA pairs in AVQA \cite{Li_2021_ICCV}.}
    \label{fig:AVQA}
\end{figure}

\begin{figure*}
    \centering
    \includegraphics[scale=0.3]{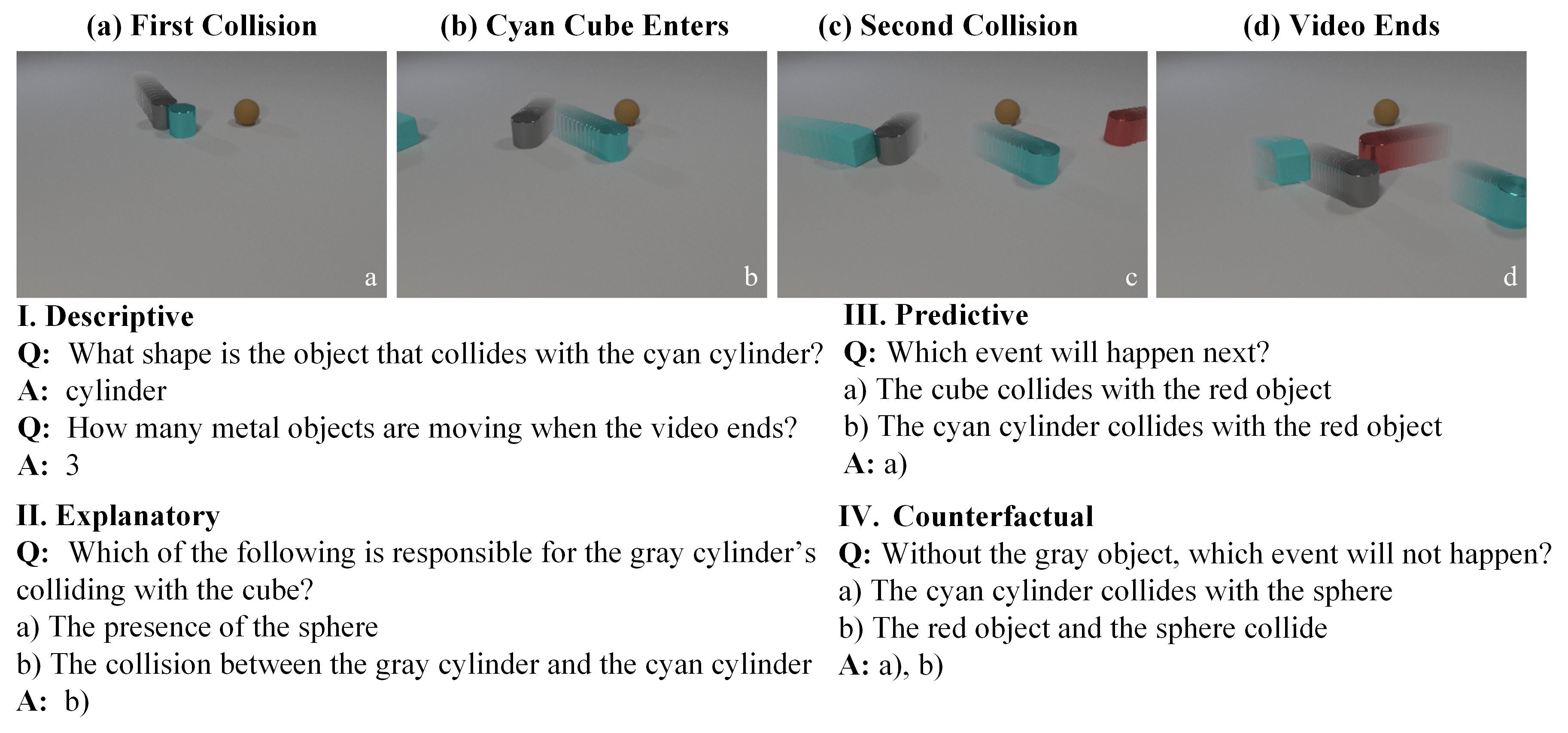}
    \caption{A sample in CLEVRER, including four question types: descriptive, explanatory, predictive, and counterfactual \cite{yi2019clevrer}.}
    \label{fig:clevrer}
\end{figure*}

\begin{figure*}
    \centering
    \includegraphics[scale=0.28]{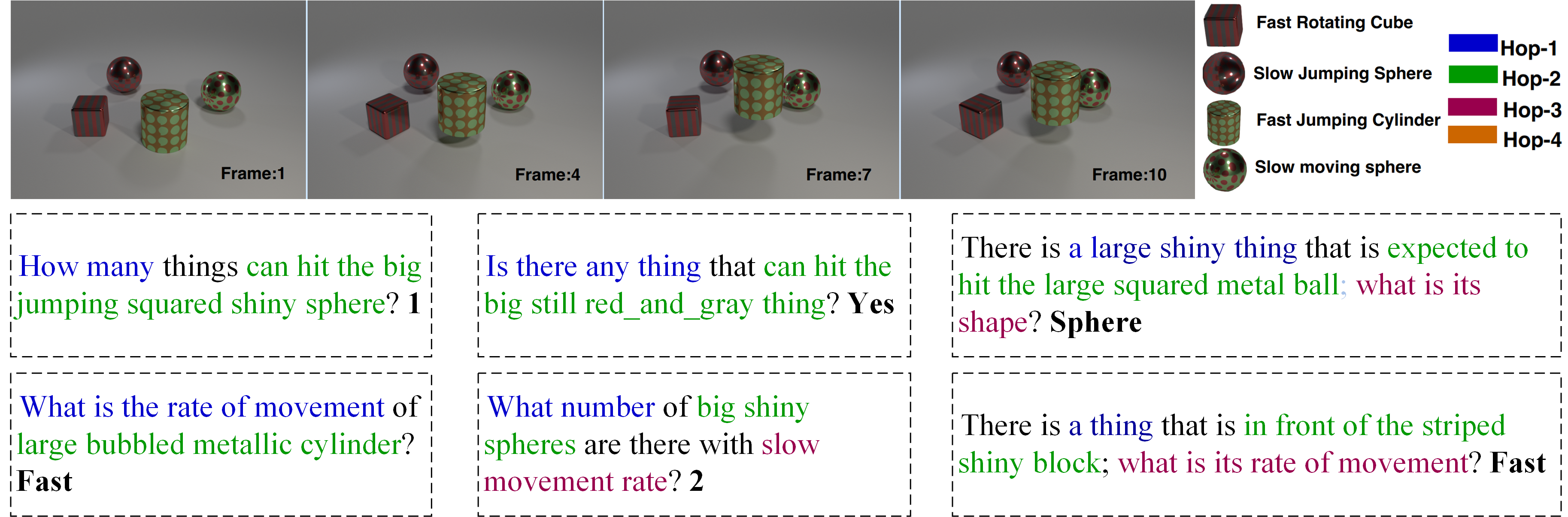}
    \caption{ Illustration of an instance of VQuAD dataset \cite{gupta2022vquad}, which shows various questions that are generated concerning the video created and the difference in complexity in terms of hops for the questions.}
    \label{fig:vqaud}
\end{figure*}

\subsection{Image Question Answering}

Image question answering benchmarks evaluate the models' capability to answering natural language questions based on a corresponding image. Recent image question answering benchmarks try to collect or generate balanced QA pairs to make the dataset distribution more balanced in question distribution. VQA v2.0 \cite{goyal2017making} collects complementary QA pairs by replacing the image and the answer in QA pairs.  VQA-CP \cite{agrawal2018don} resplited the VQA v1 dataset and VQA v2 dataset to construct two new datasets VQA-CP v1 and VQA-CP v2. As Fig. \ref{fig:IV-VQA} shows,  \citet{agarwal2020towards} constructed IV-VQA and CV-VQA datasets using semantic editing to generate images and reexamine the image by a human. \citet{Li_2021_ICCV} proposed a human-machine adversarial to collect robust QA pairs. Fig. \ref{fig:AVQA} illustrates the adversarial data collection procedure. In Table \ref{tab:vqadataset}, we summarize these datasets in terms of image source, split numbers, collected or not, and rebalanced or not.

Current image question answering uses various approaches to overcome the bias introduced by unbalanced data. However, there is still a lack of large-scale benchmark datasets that support fair and transparent evaluations of the causality behind the data and the reasoning ability of the method. Introducing the causal concept and causal methods like confounders and causal interventions when building benchmark datasets may help resolve the problem of the lack of specific causal reasoning benchmark datasets.

\begin{figure*}[t]
    \centering
    \includegraphics[scale=0.24]{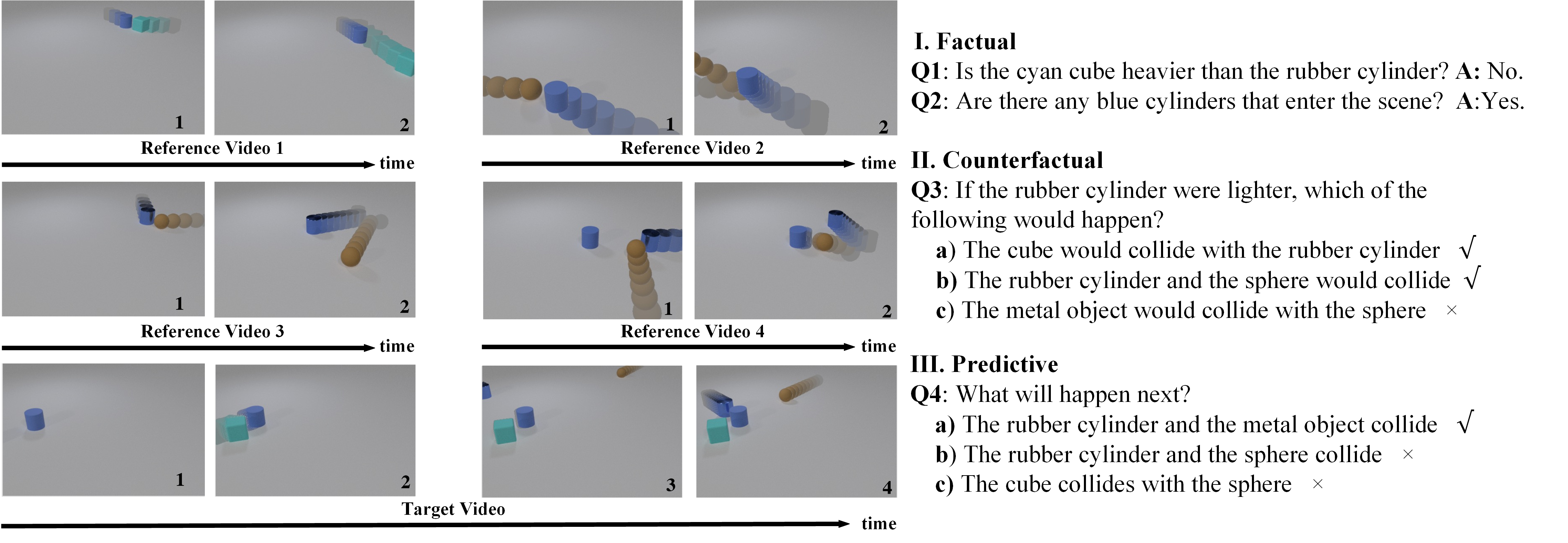}
    \caption{Sample reference videos, target video, and question-answer pairs from ComPhy dataset \cite{chen2022comphy}.}
    \label{fig:comphy}
\end{figure*}

\subsection{Video Question Answering}
The video question answering task is more complex than the image question answering task due to the ubiquitous correlation between spatial and temporal information, i.e., the introduction of complex temporal relations. Thus, improving the spatial-temporal causal reasoning ability of models can improve the performance on this task, but simply approximating data distributions usually does not work. Thus, some recently released benchmark datasets are proposed to evaluate whether the model has the reasoning ability to understand the causal relation knowledge within the visual and linguistic content, as shown in Table \ref{tab:videodatasets}.

\begin{figure}
    \centering
    \includegraphics[width = 0.43\textwidth]{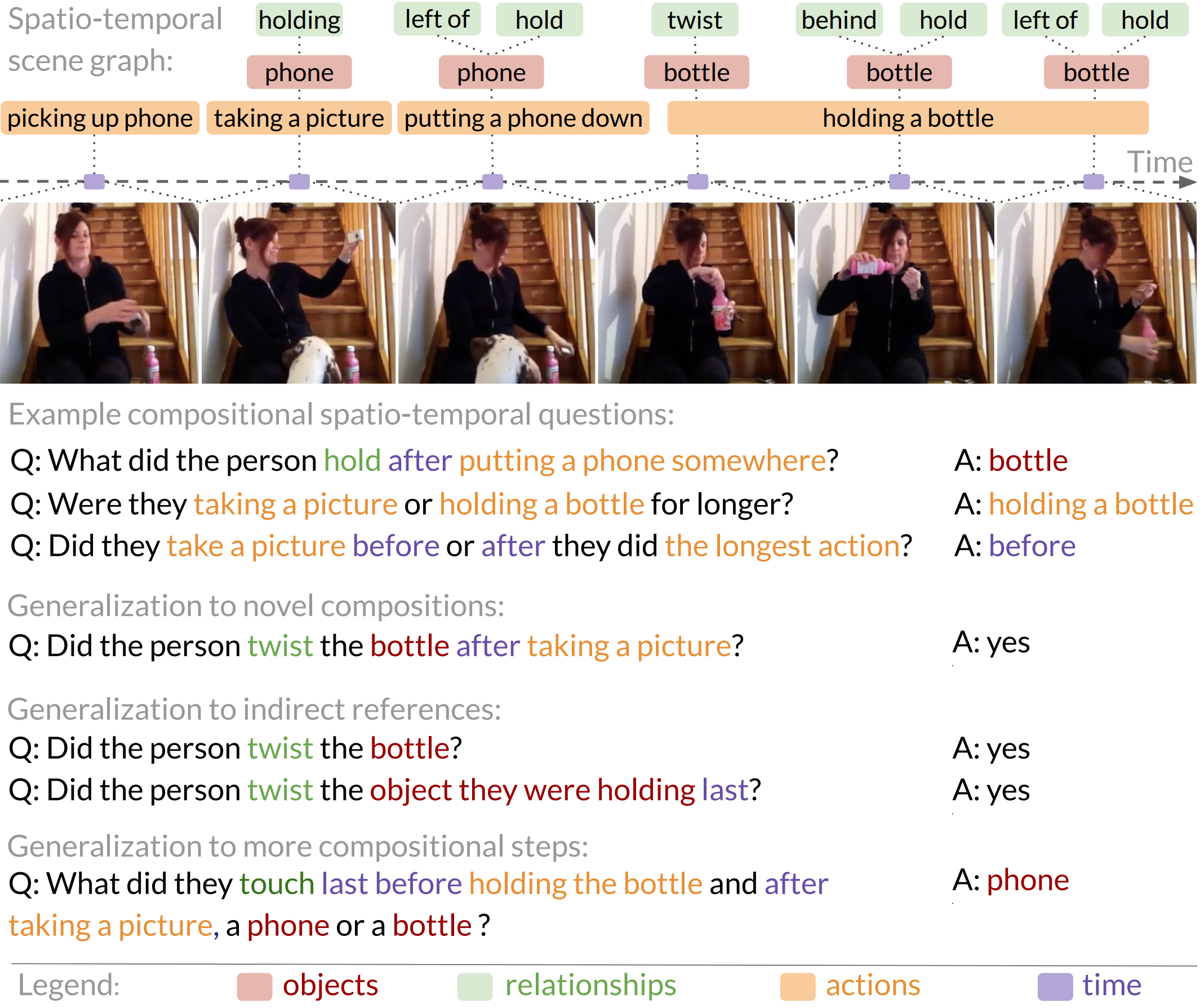}
    \caption{An overview of AGQA dataset \cite{grunde2021agqa}.}
    \label{fig:agqa}
\end{figure}

\begin{figure}
    \centering
    \includegraphics[width=0.5\textwidth]{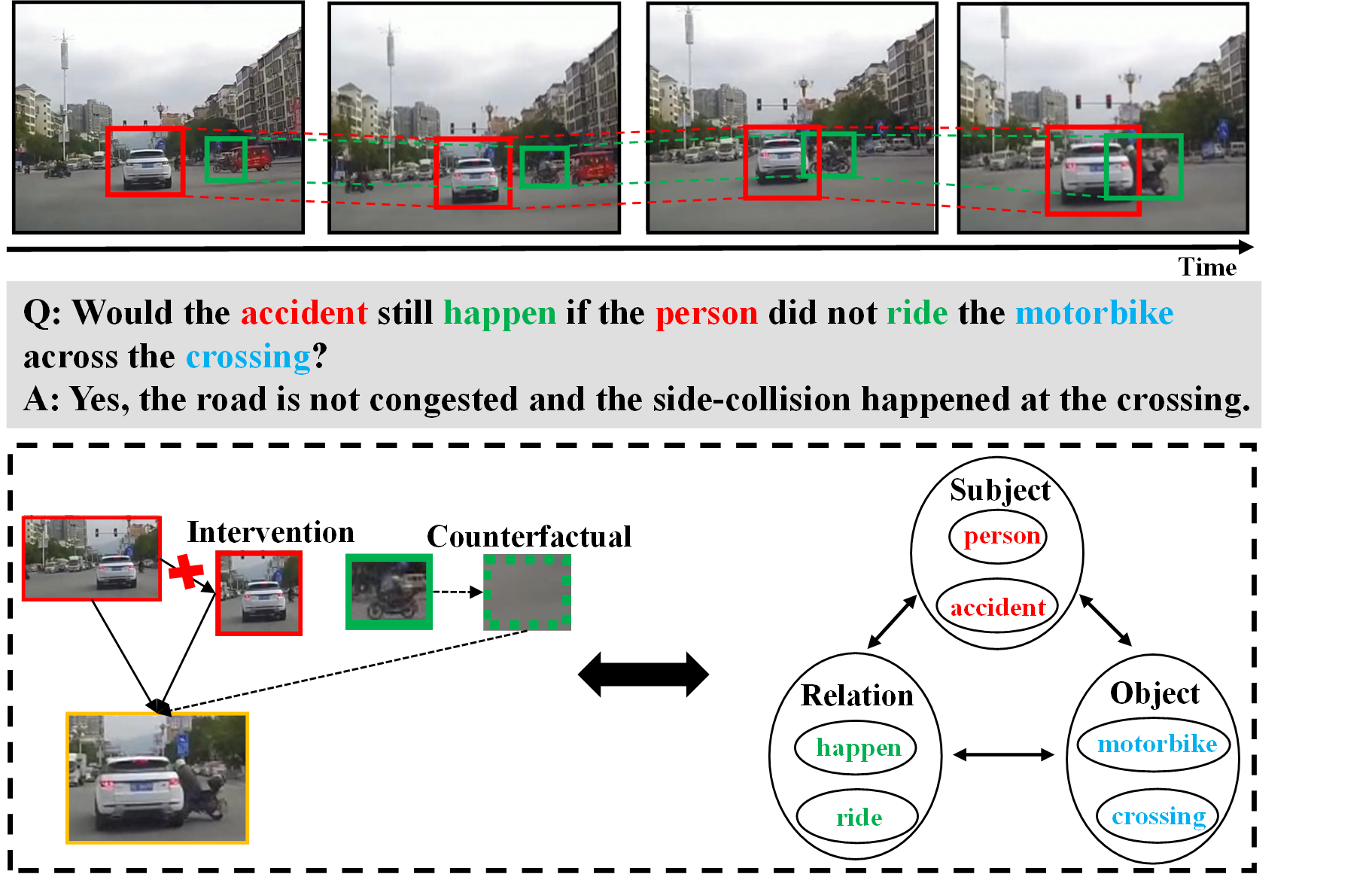}
    \caption{An example of counterfactual question-answer pair in SUTD-TrafficQA dataset \cite{xu2021sutd}.}
    \label{fig:sutd}
\end{figure}

\begin{figure*}[t]
    \centering
    \includegraphics[scale=0.5]{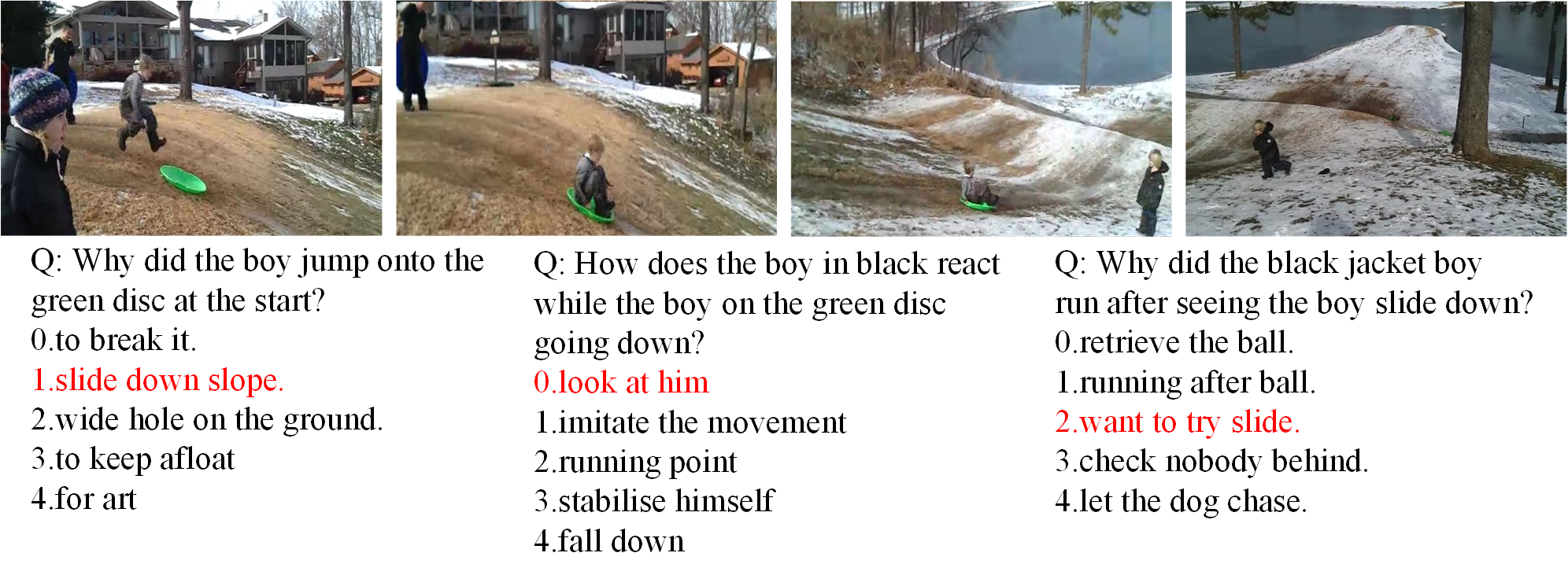}
    \caption{Examples of multi-choice QA in NExT-QA dataset \cite{xiao2021next}.}
    \label{fig:nextqa}
\end{figure*}

\begin{figure*}[t]
    \centering
    \includegraphics[scale=0.45]{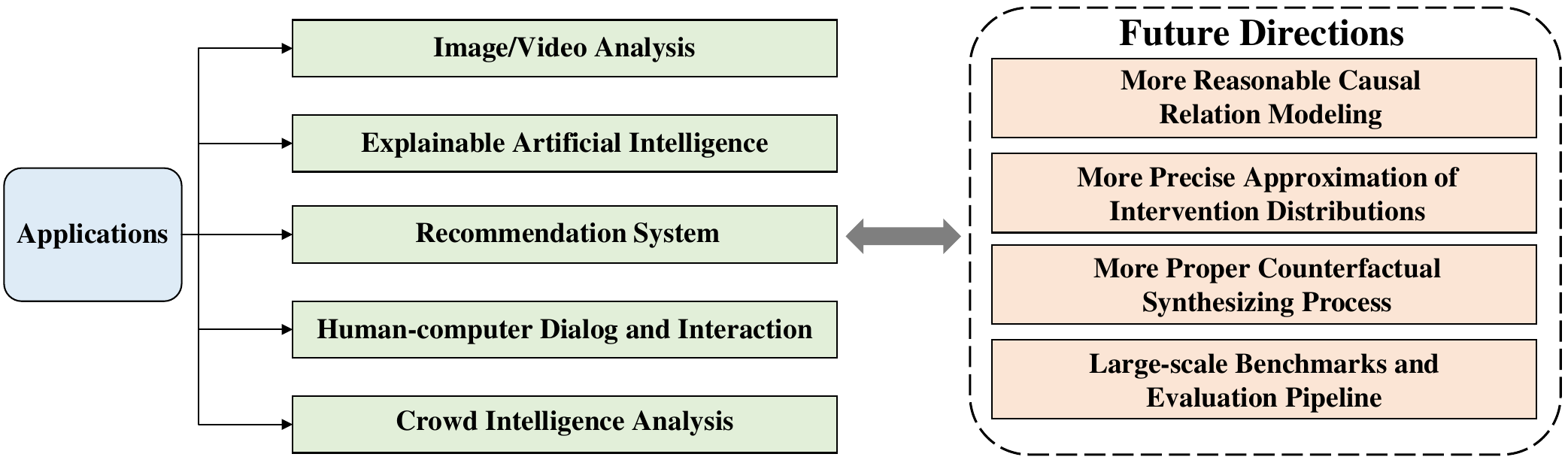}
    \caption{Extensive applications and future directions.}
    \label{fig:application}
\end{figure*}

CLEVRER \cite{yi2019clevrer} contains synthesized videos and automatically generated questions describing the collision of geometric objects. A typical video and question types from CLEVRER are shown in Fig. \ref{fig:clevrer}. It is a balanced and synthetic dataset that contains diagnostic annotations and counterfactuals. VQuAD \cite{gupta2022vquad} is also a diagnostic synthesized dataset. It is constructed from a balanced dataset by separating objects into attributes like texture and color, and balancing the data distribution based on these attributes. A brief overview of the VQuAD
objects is shown in Fig. \ref{fig:vqaud}. The VQuAD is a diagnostic dataset that can be used to evaluate the extent of reasoning abilities of various video QA methods. ComPhy \cite{chen2022comphy} is a video QA dataset that focuses on understanding object-centric and relational physics properties hidden from visual appearances. As shown in Fig. \ref{fig:comphy}, the ComPhy dataset studies object's intrinsic physical properties from object's interactions and how these properties affect their motions in future and counterfactual scenes to answer the corresponding questions. AGQA \cite{grunde2021agqa} includes numerous QA pairs, which are automatically generated by the process. An overview of AGQA is shown in Fig. \ref{fig:agqa}. The QA pairs in AGQA are generated by parsing videos to scene graphs and using the language composition inference by scene graph to generate QA pairs. SUTD-TrafficQA \cite{xu2021sutd} is a traffic video question answering dataset with six challenging reasoning tasks, including basic understanding, event forecasting, reverse reasoning, counterfactual inference, introspection, and attribution analysis, to analyze the models' reasoning ability. Fig. \ref{fig:sutd} shows an example of a counterfactual traffic video question answering process from SUTD-TrafficQA.  To be noticed, the counterfactual traffic video question answering task in Fig. \ref{fig:sutd} requires the outcome of a certain hypothesis that does not occur in the video. To accurately reason about the imagined events under the designated condition, the model is required to not only conduct relational reasoning in a hierarchical way but also fully explore the causal, logic, and spatial-temporal structures of the visual and linguistic content. NExT-QA \cite{xiao2021next} is a video question answering benchmark targeting the explanation of the content of the video, which requires a deeper understanding of videos and reasoning about causal and temporal actions from rich object interactions in daily activities. As shown in Fig. \ref{fig:nextqa}, the NExT-QA dataset contains rich object interactions and requires causal and temporal action reasoning in realistic videos. The NExT-QA dataset challenges QA models to reason about causal and temporal actions and understands rich object interactions in daily activities.

\section{Extensive Applications}
Causal reasoning with visual representation learning has a variety of applications. Modeling causal reasoning for a variety of tasks can achieve a better perception of the real world. In this section, we introduce the applications from five aspects: image/video analysis, explainable artificial intelligence, recommendation system, human-computer dialog and interaction, and crowd intelligence analysis. We also discuss how causal reasoning benefits various real-world applications, as shown in Fig. \ref{fig:application}.

In image/video analysis, most of the existing work relies on learning data correlations rather than causal structures, and the superficial correlation within the image and video data makes the model vulnerable to visual changes in the dataset. Therefore, a causality-ware feature learning strategy is required to make the model learn essential causal structures behind the data and robust to different data distributions. One of the main methods of dealing with superficial data correlations is using the causal intervention. Assume that commonsense knowledge exists in visual features, but commonsense might be confused by false observation bias. For example, the words ``cup", ``table", and ``stool" have high co-occurrence frequencies because they commonly appear in daily life, but the commonsense knowledge usually wrongly predicts the class as table due to the observation bias. To reduce the observation bias, the causality-ware visual commonsense model is required, which regards the object category as a confounding factor and directly maximizes the likelihood after the intervention to learn the visual feature representation. By eliminating observation bias, the learned visual features are robust to image and video analysis tasks. A representative task is weakly supervised object localization, detection, and grounding \cite{zhang2020weakly, zhang2021weakly,wang2022weakly}, which aims to localize objects described in the sentence to visual regions in the image/video. Despite recent progress, existing methods may suffer from the severe problem of spurious association such as: (1) the association is not object-relevant but extremely ambiguous due to weak supervision, and (2) the association is unavoidably confounded by the observational bias when using statistics-based methods. Therefore, a unified causal framework is required to learn the deconfounded object-relevant association for accurate and robust video object localization, detection and grounding.

With the development of deep learning across industries and disciplines, the applications of deep learning models in real-world scenes require a high degree of robustness, interpretability, and transparency. Unfortunately, the black-box properties of deep neural networks are still not fully explainable, and many machine decisions are still poorly understood \cite{tjoa2020survey}. In recent years, causal interpretability has received increasing attention. These works \cite{parafita2019explaining, narendra2018explaining, harradon2018causal, chattopadhyay2019neural,moraffah2020causal,o2020generative,lin2021generative} have made progress in explainable artificial intelligence based on causal interpretability. For example, in the current COVID-19 pandemic, causal mediation analysis helps disentangle different effects contributing to case fatality rates when an example of Simpson's paradox was observed \cite{von2021simpson}.
Learning the best treatment rules for each patient is one of the promising goals of applying explainable treatment effect estimation methods in the medical field. Since the effects of different available drugs can be estimated and explainable, doctors can prescribe better drugs accordingly.

At present, some causal reasoning works \cite{zheng2021disentangling,liu2021mitigating,wei2021model,wang2021clicks,zhang2021causal} have been applied to the recommendation system. The recommendation system is actually a problem of causal reasoning \cite{zheng2021disentangling}. User embedding represents what type of person the user is and infers the user's preferences based on the user's attributes. The causal effect of a recommendation system is whether the user is satisfied with the recommendation. Superficial bias exists because the recommendation system is trained on biased samples (both users and items). An example is a personalized recommendation, where we wish to build a model of a customer's shopping interest through various data sources, such as webbrowser records and shopping history. However, if we train a recommendation system on customers' records in controlled settings, the system may provide little additional insight compared to the customers' mental states and  emotions, thus may fail when deployed. While it may be useful to automate certain decisions, understanding causality may be necessary to recommend commodities that are personalized and reliable. A general approach to removing survival bias is to construct counterfactual mirror users, construct similarity measures using unbiased information, and construct matches from low-active to high-active users. In this way, we can alleviate the user's dissatisfaction with the previously recommended content and the low user activity.

For human-computer dialog and interaction, some emerging tasks contain the interaction between vision and language. Additionally, there exist multi-modal spatial-temporal information and complex relations captured by various devices.  Most of the existing work relies on data correlation rather than causal relevant evidence, and the false correlation in the data makes the model vulnerable to language biases in the problems. Take a VQA task as an example, where we aim to remove visual objects that are unrelated to answering the question, and the prediction of the model is not expected to change. This can prevent the model from relying on superficial data correlations. When changing objects that are related to a question, the model is expected to change the answer accordingly. Adjusting question-related objects encourages the model to predict based on causality-aware objects. For a better user experience, the human-computer dialog and interaction system is required to understand people's purposes and make reliable decisions. Causal reasoning is beneficial to the pursuit of reliable human-computer interaction by uncovering and modeling heterogeneous spatial-temporal information in a reliable and explainable way. Especially for robot interaction \cite{xiong2016robot, stocking2022robot, lee2021causal, smith2020counterfactual}, where the relevant environmental features are not known in advance, prior knowledge can be utilized as a good candidate for causal structures. The strong relation between causal reasoning and its ability to intervene in the world suggests that causal reasoning can greatly address this challenge for robotics, which benefits the application of robotics significantly.

The applications mentioned above usually focus on a single subject, whereas crowd intelligence analysis \cite{hou2019mobile} aims to address related sensing and cognitive tasks for multiple subjects and their interaction. In recent years, we have been witnessing the explosive growth of multi-modal heterogeneous spatial/temporal/spatial-temporal data from different kinds of data sensors. Urban computing \cite{zheng2014urban} is an example of crowd intelligence analysis, which aims to tackle traffic congestion \cite{TFP}, energy consumption \cite{zhu2022hybrid}, and pollution by using the data that has been generated by a large number of traffic vehicles in cities (e.g., traffic flow, human mobility, and geographical data). For example, huge amounts of heterogeneous traffic data come from various sources, including both static and dynamic data, such as traffic road networks, geographic information system (GIS) data, traffic flow, traffic mobility, traffic energy consumption, etc. Moreover, the heterogeneous spatial-temporal traffic data contains a large number of useful traffic rules with strong causal relations. Therefore, how to utilize different heterogeneous spatial-temporal data and discover their complex and entangled causal relations is beneficial to urban computing and crowd intelligence analysis.

\section{More Detailed Discussions}

Some researchers have successfully implemented causal reasoning for visual representation learning to discover causality and visual relations. However, causal reasoning for visual representation learning is still in its infancy stage, and many issues remain unsolved. Therefore, this section highlights several possible research directions and open problems to inspire further extensive and in-depth research on this topic. Potential research directions for causal visual representation learning can be summarized as: 1) More reasonable causal relation modeling; 2) More precise approximation of intervention distributions; 3) More proper counterfactual synthesizing process; 4) Large-scale benchmarks and evaluation pipeline.

\subsection{More Reasonable Causal Relation Modeling}
Reasonable causality modeling is the basis for causal inference. Real-world data like visual information is usually unstructured, and the effect of causal relation may be unobserved. For example, momentum is likely to be detrimental under long-tailed distribution data \cite{tang2020long}, and there is no consensus on how to properly model causality on many tasks because the real causality may be more complicated than expected. For the VQA task, \citet{yang2021causal} treated visual and language features as one vertex in the causal graph, and \citet{niu2021counterfactual} consider the visual and linguistic features separately. However, these methods focus on causality discovery on either visual or linguistic modality without considering both of them. Therefore, future work should consider: 1) In-depth analysis of the relations between causal reasoning and visual representation learning; 2) Model comprehensive and reasonable causal relation.

\subsection{More Precise Approximation of Intervention Distributions}
A precise estimation of the intervention distribution helps the implementation of a certain causal model. Most of the current intervention distribution approximation methods focus on identifying all confounders for a certain task, while these confounders are usually defined as the average of object features in visual tasks \cite{wang2020visual, wang2021causal, zhang2021devlbert}. Actually, the average features may not properly describe a certain confounder, especially for complex heterogeneous visual data. Thus, how to approximate the confounders more accurately is a key future work that needs to be further considered for causal intervention methods.

\subsection{More Proper Counterfactual Synthesising Process}
Counterfactual inference-based methods usually focus on refining the training procedure, i.e., embedding the counterfactual inference process into the training procedure. Counterfactual synthesizing methods \cite{agarwal2020towards, chen2020counterfactual, yue2021counterfactual, mao2021generative} have proved their effectiveness in many tasks. Embedding counterfactual inference into models can effectively eliminate data bias within the data. A novel counterfactual framework \cite{niu2021counterfactual} gives us insight into this potential. However, visual data is often entangled and heterogeneous, which makes the data bias hard to understand and model. Therefore, how to model a proper counterfactual synthesizing process is a potential direction of data debiasing in visual representation.

\subsection{Large-scale Benchmarks and Evaluation Pipeline}

Although causal reasoning has been burgeoned for many visual learning tasks, most of the existing evaluation datasets are still traditional datasets for correlation learning without proper large-scale benchmarking datasets and pipelines to support fair and transparent evaluations of emerging research contributions. The only existing causal datasets discussed in the above sections have limited scale and lack comprehensive evaluation standards for causal reasoning. Therefore, more large-scale benchmark datasets and pipelines for specific visual representation learning tasks should be considered in future research.

Generally, causal visual representation learning is still an emerging and challenging research topic. Causal modeling, intervened distribution approximation, counterfactual inference, large-scale benchmarks, and evaluation pipelines have great potential for further exploration.


\section{Conclusions}\label{sec13}
This paper has provided a comprehensive survey on causal reasoning for visual representation learning. In this paper, we focus on the prospective survey of related works, datasets, insights,
future challenges and opportunities for causal reasoning, visual representation learning, and their integration. We mathematically present the basic concepts of causality, the structural causal model (SCM), the independent causal mechanism (ICM) principle, causal inference, and causal intervention. Then, based on the analysis, we further give some directions for conducting causal reasoning on visual representation learning tasks. We also review some recent popular visual learning tasks, including visual understanding, action detection and recognition, and visual question answering, including the discussions about the existing challenges of these visual learning methods. In addition, the related causality-based visual representation learning works and datasets are also discussed systematically. Finally, extensive applications and some potential future research directions are provided for further exploration. We hope that this survey can help attract attention, encourage discussions, and bring to the forefront the urgency of developing novel causal reasoning methods, publicly available benchmarks, and consensus-building standards for reliable visual representation learning and related real-world applications more efficiently.









\section*{Acknowledgments}
This work was supported in part by the National Natural Science Foundation of China under Grant No.62002395, No.61976250, and No.U1811463, in part by the National Key R\&D Program of China under Grant No.2021ZD0111601, in part by the Guangdong Basic and Applied Basic Research Foundation under Grant No.2021A15150123 and No.2020B1515020048.

\bibliographystyle{IEEEtranN}
\bibliography{bibfile}



\begin{figure}[h]%
\centering
\includegraphics[width=0.1\textwidth]{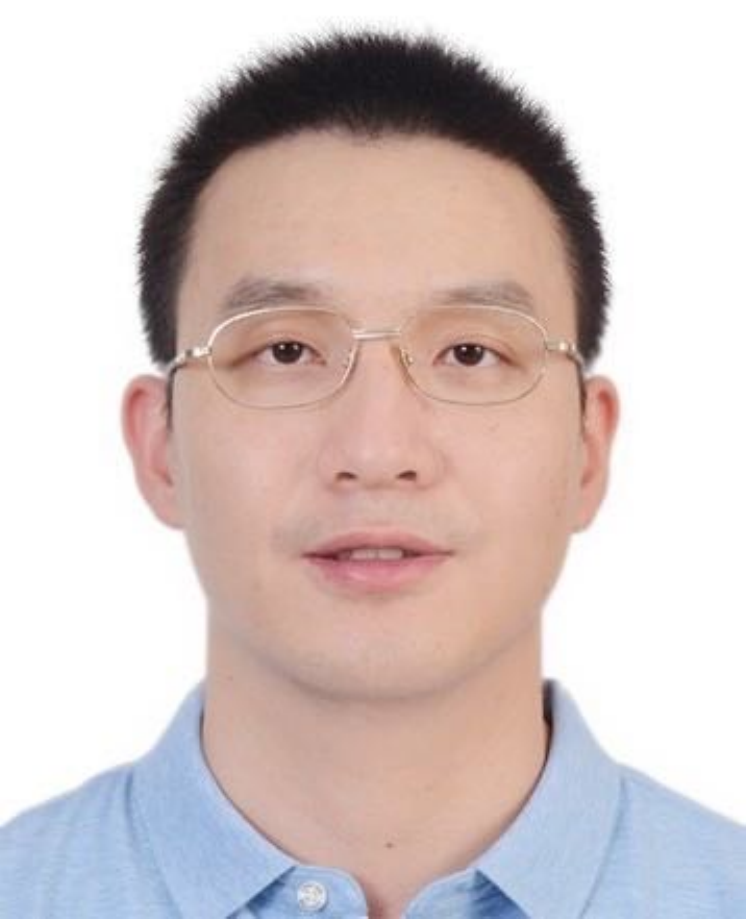}
\end{figure}
\noindent{\bf Yang Liu}\quad is currently a research associate professor working at the School of Computer Science and Engineering, Sun Yat-sen University. He received the B.Sc. degree in telecommunications engineering from Chang’an University, Xi’an, China, in 2014, and the Ph.D. degree in telecommunications and information systems from Xidian University, Xi’an, in 2019. He has authorized and co-authorized more than 20 papers in top-tier academic journals and conferences. He has been serving as a reviewer for numerous academic journals and conferences such as IEEE TIP, TNNLS, TMM, TCSVT, TCyb, CVPR, ICCV, AAAI, and ECCV. He is a member of IEEE and CSIG. More information can be found on his homepage https://yangliu9208.github.io/home.

His current research interests include video understanding, causal reasoning, and computer vision.

E-mail: liuy856@mail.sysu.edu.cn

ORCID iD: 0000-0002-9423-9252


\begin{figure}[h]%
\centering
\includegraphics[width=0.1\textwidth]{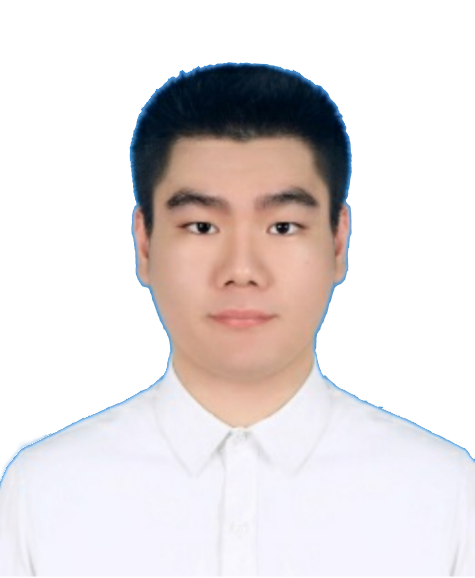}
\end{figure}
\noindent{\bf Yushen Wei} received the B.S. degree in computer science and technology from Sun Yat-sen University, Guangzhou, China, in 2020. He is currently pursuing the master’s degree with the School of Computer Science and Engineering, Sun Yat-sen University.

His current research interests include video understanding, computer vision and machine learning.

E-mail: weiysh8@mail2.sysu.edu.cn

ORCID iD: 0000-0002-0527-5463

\begin{figure}[h]%
\centering
\includegraphics[width=0.1\textwidth]{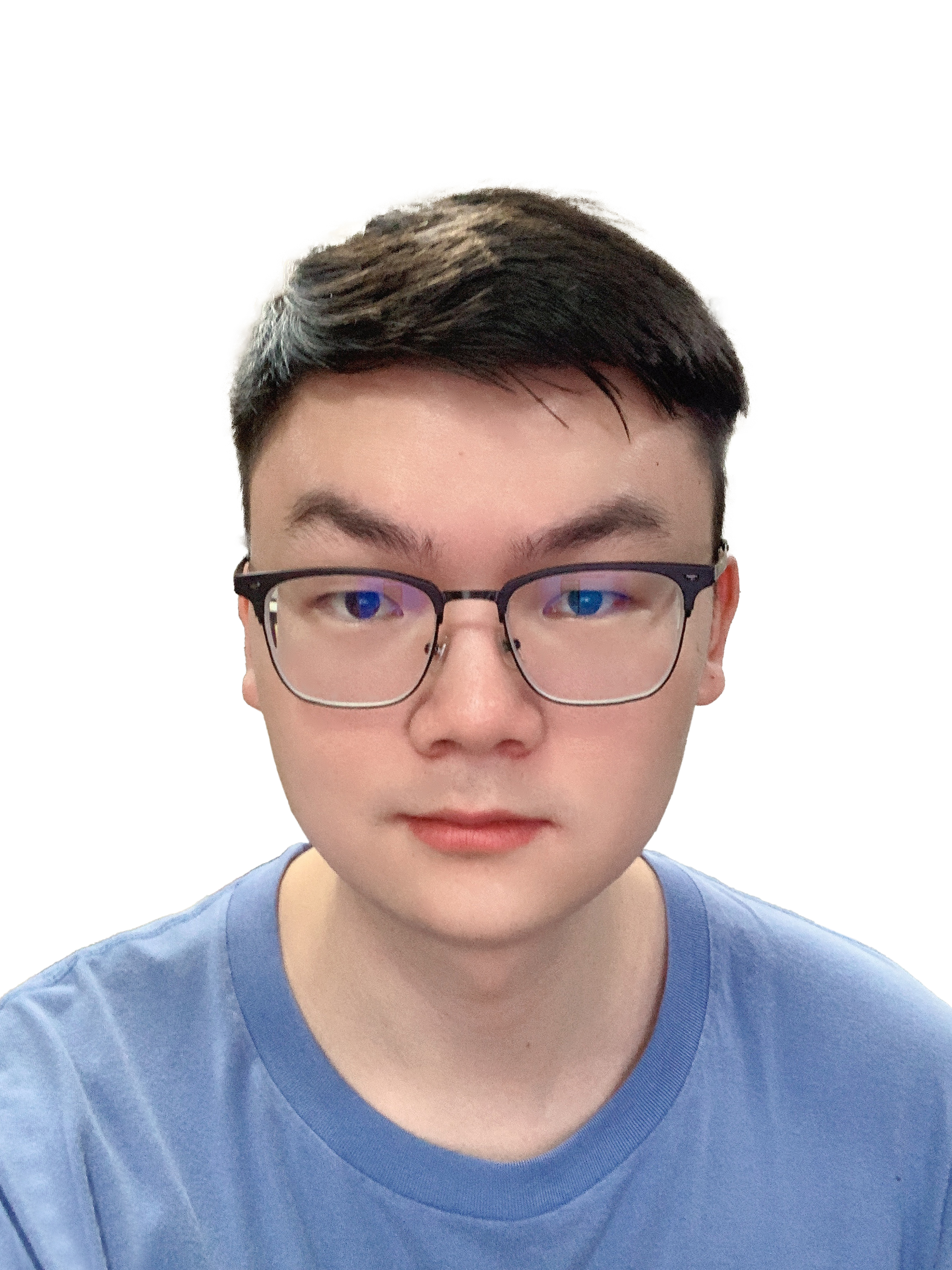}
\end{figure}
\noindent{\bf Hong Yan} received the B.S. degree in computer science and technology from Nanchang University, Nanchang, China, in 2020. He is currently pursuing the master’s degree with the School of Computer Science and Engineering, Sun Yat-sen University.

His current research interests include video understanding, computer vision and machine learning.

E-mail: yanh36@mail2.sysu.edu.cn

ORCID iD: 0000-0003-4100-6751

\begin{figure}[h]%
\centering
\includegraphics[width=0.1\textwidth]{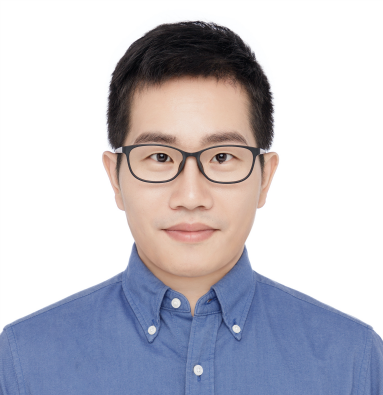}
\end{figure}

\noindent{\bf Guanbin Li} is currently an associate professor in School of Computer Science and Engineering, Sun Yat-Sen University. He received his PhD degree from the University of Hong Kong in 2016. His current research interests include computer vision, image processing, and deep learning. He is a recipient of ICCV 2019 Best Paper Nomination Award. He has authorized and co-authorized on more than 70 papers in top-tier academic journals and conferences. He serves as an area chair for the conference of VISAPP. He has been serving as a reviewer for numerous academic journals and conferences such as TPAMI, IJCV, TIP, TMM, TCyb, CVPR, ICCV, ECCV and NeurIPS.

His research interests include computer vision and machine learning.

E-mail: liguanbin@mail.sysu.edu.cn

ORCID iD: 0000-0002-4805-0926

\begin{figure}[h]%
\centering
\includegraphics[width=0.1\textwidth]{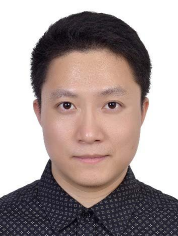}
\end{figure}

\noindent{\bf Liang Lin} is a Full Professor of computer science at Sun Yat-sen University. He served as the Executive Director and Distinguished Scientist of SenseTime Group from 2016 to 2018, leading the R\&D teams for cutting-edge technology transferring. He has authored or co-authored more than 200 papers in leading academic journals and conferences, and his papers have been cited by more than 21,000 times. He is an associate editor of IEEE Trans. Neural Networks and Learning Systems and IEEE Trans. Human-Machine Systems, and served as Area Chairs for numerous conferences such as CVPR, ICCV, SIGKDD and AAAI. He is the recipient of numerous awards and honors including Wu Wen-Jun Artificial Intelligence Award, the First Prize of China Society of Image and Graphics, ICCV Best Paper Nomination in 2019, Annual Best Paper Award by Pattern Recognition (Elsevier) in 2018, Best Paper Dimond Award in IEEE ICME 2017, Google Faculty Award in 2012. His supervised PhD students received ACM China Doctoral Dissertation Award, CCF Best Doctoral Dissertation and CAAI Best Doctoral Dissertation. He is a Fellow of IET/IAPR.

His research interests include artificial intelligence, computer vision, machine learning, multimedia, and NLP/Dialogue.

E-mail: linliang@ieee.org (Corresponding author)

ORCID iD: 0000-0003-2248-3755

\end{document}